\documentclass{article}

\usepackage{iclr2027_conference,times}
\usepackage{amsmath,amssymb}
\usepackage{booktabs}
\usepackage{graphicx}
\usepackage{microtype}
\usepackage{float}
\usepackage{hyperref}
\usepackage{url}
\usepackage{wrapfig}
\usepackage{graphicx}

\usepackage{wrapfig}
\usepackage{graphicx}
\usepackage{booktabs}
\usepackage{enumitem}
\usepackage{array}

\usepackage{booktabs}
\usepackage{pdflscape}
\usepackage{longtable}

\graphicspath{{figures/}}

\vspace{-20pt}
\title{Compact Robot Policies Need Fine-Grained\\Visual Representations}

\iclrfinalcopy
\author{
Nanhe Chen\textsuperscript{$\dagger$}
\hspace{1.1em}
Runqiu Yang\textsuperscript{$\dagger$}
\hspace{1.1em}
Jiawei Tang\textsuperscript{$\dagger$}
\hspace{1.1em}
Sichao Liu\textsuperscript{$\ddagger \ast$}
\hspace{1.1em}
Yuquan Wang\textsuperscript{$\dagger \ast$}
\\[1.5ex]
$\dagger$~X Square Robot \quad
$\ddagger$~KTH
}

\ifdefined\ICLRFINAL
  \iclrfinalcopy
\fi

\begin{document}

\maketitle
\begingroup
\renewcommand\thefootnote{}
\footnotetext{$\ast$ Corresponding author: Yuqan Wang (\url{haas@x2robot.com}) \& Sichao Liu (\url{sicliu@kth.se})}
\endgroup
\fancyhead[L]{Preprint}

\begin{abstract}
Multi-task manipulation policies differ in architecture, scale, and pretrained priors all at once, so published comparisons cannot attribute performance to any single component. 
\textbf{We argue that most of it comes from the visual representation, and that parameter scale and generative priors are largely incidental}. 
To test this, we build \textbf{CoRP} (Compressed Representation Policy), a deliberately compact policy (\textbf{48.9M parameters, no vision-language model and no video-generative prior}) that factorizes into a representation extractor and a flow-matching action generator. 
It reaches \textbf{97.0\% on LIBERO} and \textbf{75.78\%/73.36\% on RoboTwin~2.0 Clean/Randomized}, matching systems 40.9–163.6× larger. 
Holding the action generator fixed, we then vary one extractor property at a time. 
\textbf{Pretrained initialization is decisive}: a random ViT-S/14 drops to 78.1\%  and an ImageNet ResNet-34 to 74.5\% on LIBERO. 
\textbf{Pretraining alone is not enough, as freezing the encoder costs 19.8 points. Compression matters as much}: resampling each view to 48 tokens beats passing all patch tokens (\textbf{97.0\% vs 83.2\%}), and a variational information bottleneck over those tokens is worse than a hard token budget, cutting LIBERO-Goal from \textbf{95.8\% to 33.0\%} by suppressing the instruction-dependent token selection the policy relies on. 
\textbf{Language conditioning contributes only where the observation leaves the goal ambiguous} (LIBERO-Goal: 9.2\% to 95.8\%), while on RoboTwin~2.0, where observations are unambiguous, removing it slightly improves success. 
Therefore, we argue that a compact policy works when its representation is pretrained, task-adapted, and compressed. Project page: \url{https://corp-policy.github.io/}
\end{abstract}

\section{Introduction}
\label{sec:introduction}
Vision-language-action models (VLAs) have become the dominant paradigm for multi-task robotic manipulation, mapping images and language instructions directly to robot actions~\citep{zitkovich2023rt2,kim2024openvla,black2025pi0}.
These systems typically couple a pretrained vision–language model (VLM) with a generative action head, carrying billions of parameters to generalize across tasks~\citep{kim2024openvla,black2025pi0}.
More recently, world-action models (WAMs) go further, adding video-generative priors that jointly predict future observations and robot actions~\citep{ye2026worldactionmodels}. Because these systems differ in architecture, scale, pretraining, data, and training recipes, we cannot attribute their success to any single component.

We factor a manipulation policy into representation extraction and conditional action generation:
\begin{equation}
\mathbf{z}_t = f_\phi(\mathbf{o}_t, \mathbf{s}_t, \ell), \qquad
\mathbf{a}_{t:t+H-1} \sim p_\theta(\,\cdot \mid \mathbf{z}_t),
\label{eq:decomposition}
\end{equation}
where $t$ is the control step, the representation extractor $f_\phi$ (parameters $\phi$) maps the observation $\mathbf{o}_t$, state $\mathbf{s}_t$, and instruction $\ell$ to the conditioning representation $\mathbf{z}_t$, and the action generator $p_\theta$ (parameters $\theta$), a DiT-style model~\citep{peebles2023scalable} trained with flow matching~\citep{lipman2023flowmatching}, samples an action chunk $\mathbf{a}_{t:t+H-1}$ of $H$ steps. The two lines pursue different kinds of generalization: VLAs seek semantic and task generalization through pretrained vision–language backbones and heterogeneous co-training~\citep{kim2024openvla,black2025pi05}, while WAMs seek temporal and physical generalization through pretrained video priors~\citep{pai2025mimicvideo,ye2026worldactionmodels}.
Yet both ultimately rely on stronger pretrained representations, and neither isolates how much of its performance the representation alone accounts for. \emph{We argue that, in multi-task manipulation, most of this performance comes from the visual representation, while parameter scale and generative priors are largely incidental.} We therefore ask a narrower question: how much of a policy's performance depends on its representation, and which of its properties matter? To answer it, we hold the action generator fixed and vary one property of $f_\phi$ at a time: how the visual encoder is initialized, whether it is adapted during training, how many visual tokens reach the action generator, and whether the instruction is available.

We instantiate the factorization as the~\textbf{Co}mpressed~\textbf{R}epresentation~\textbf{P}olicy (\textbf{CoRP}), a deliberately compact and modular policy.
CoRP has only 48.9M parameters and four standard components: a DINOv2-initialized ViT-S/14 visual encoder~\citep{oquab2024dinov2}, a T5-Efficient-Tiny language encoder with a lightweight adapter~\citep{raffel2020t5,tay2022scale}, a learned-query cross-attention resampler that compresses each camera view into 48 tokens, and a single-block DiT action generator trained with flow matching. With few interacting parts, each intervention changes exactly one factor, so we can directly attribute its effect. Despite having neither a VLM nor a video-generative prior, CoRP reaches 97.0\% average success on LIBERO~\citep{liu2023libero}, matching VLA and WAM systems 40–160× larger, and 75.8\%/73.4\% on RoboTwin~2.0 Clean/Randomized~\citep{chen2025robotwin}, surpassing $\pi_0$, $\pi_{0.5}$, and X-VLA while remaining below video-pretrained WAMs (Tables~\ref{tab:libero_comparison} and~\ref{tab:robotwin_comparison}). This result is our starting point, not our contribution: our controlled interventions locate its source.

Our interventions identify the representation extractor as the main source of this performance and show what makes it fine-grained enough for control: the representation must be \emph{pretrained, task-adapted, and compressed}. Our study is restricted to closed-set benchmarks; we do not evaluate the open-world semantic and physical generalization that VLAs and WAMs target~\citep{black2025pi05,ye2026worldactionmodels}.

We summarize our contributions as follows:
\begin{itemize}[itemsep=0pt, topsep=0pt, leftmargin=*]
    \item~\textbf{Compressed Representation Policy (CoRP)}, a 48.9M-parameter policy without any VLM or video-generative prior that is competitive with systems up to 163× larger on LIBERO and RoboTwin~2.0, and serves as a controlled testbed for isolating the role of representation.
    \item~\textbf{A controlled ablation study} shows that effective compact policies require visual representations that are pretrained, task-adapted, and compressed, and that language helps only when the goal is visually ambiguous.
    \item~\textbf{Real-robot validation} on three bimanual tasks (closed-loop adaptation to object displacement, coordinated two-arm placement, and precision plug insertion), providing qualitative evidence that CoRP, pretrained on our large-scale real-world corpus, transfers to physical hardware.
\end{itemize}

\section{Related Work}
\label{sec:related-work}

\textbf{Generative and Pretrained Policies.}
Diffusion Policy~\citep{chi2023diffusionpolicy} and ACT~\citep{zhao2023act} established generative action heads as strong baselines for visuomotor control, but each model trains on a single task. 
Octo~\citep{ghosh2024octo} scaled such policies across tasks and embodiments by pretraining on large multi-robot datasets, but learns its visual tokenizer from robot data rather than initializing it from a large-scale pretrained vision model, a choice our encoder ablation examines directly (Section~\ref{sec:exp-init}). VLAs instead build the policy on a pretrained vision–language model to inherit its richer visual and semantic representations~\citep{zitkovich2023rt2, kim2024openvla,black2025pi0,black2025pi05}. World-action models add video-generative priors to target temporal and physical generalization, either alongside a vision–language understanding module~\citep{bi2026motus} or in place of it~\citep{yuan2026fastwam}. Across both lines, however, the visual pathway changes alongside the language model, the generative prior, and the parameter count, so no single factor can be credited with the gains. In contrast, we hold a small flow-matching action generator fixed and vary only the representation extractor, testing how far a compact policy can go without either prior.

\textbf{Visual Representation Learning.}
Supervised pretraining on large labeled datasets established convolutional~\citep{he2016deep} and patch-based Transformer~\citep{dosovitskiy2021image} backbones as sources of transferable visual features, and our ImageNet ResNet-34 ablation follows this recipe.
Self-supervised methods remove the need for labels: the DINO series~\citep{caron2021dino, oquab2024dinov2, simeoni2025dinov3} yields the fine-grained patch features our resampler consumes. Recent work pushes further toward dense geometric structure, for example through masked boundary modeling~\citep{lingbot-vision2026}. Another line pretrains representations specifically for manipulation from human video, masked image modeling, or sensorimotor prediction~\citep{nair2022r3m,radosavovic2023robot,majumdar2023we}, and finds that the benefit of pretrained features varies substantially across tasks and evaluation settings~\citep{burns2023makes,hansen2022pre}. However, these studies mostly evaluate frozen features with single-task or lightweight policy heads, leaving open how initialization, fine-tuning, and token compression jointly shape a multi-task, language-conditioned generative policy. We address this gap by comparing a DINOv2-initialized ViT-S/14 against random and ImageNet-supervised alternatives, frozen and fine-tuned, under an identical action generator (Sections~\ref{sec:exp-init} and~\ref{sec:exp-adaptation}).

\section{Method}
\label{sec:method}
 
\begin{wrapfigure}{r}
{0.7\textwidth}
    \centering
    \vspace{-5pt}
    \includegraphics[width=\linewidth]{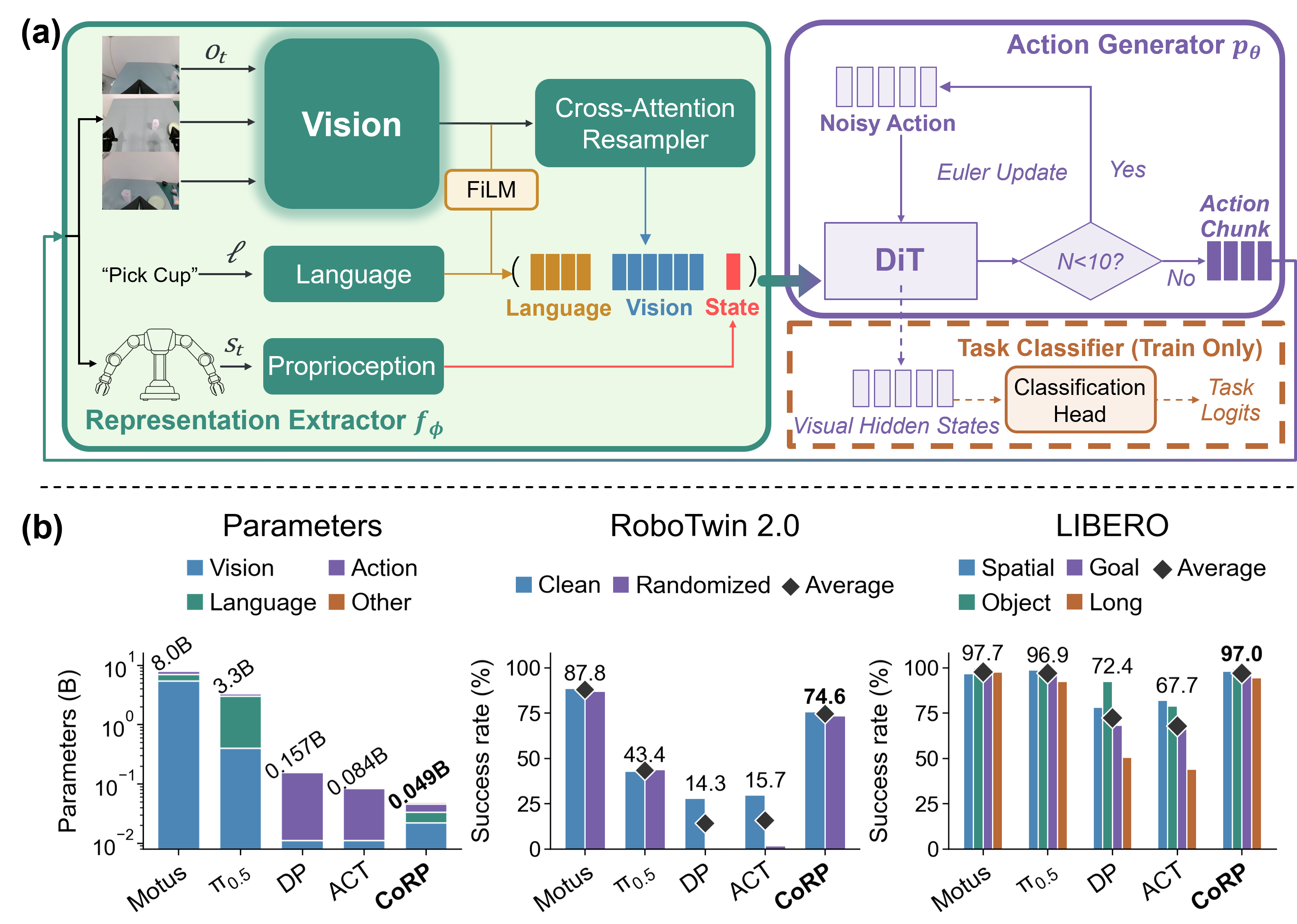}
    \vspace{-15pt}
    \caption{
    (a)~\textbf{CoRP's architecture}. The representation extractor (DINOv2, T5, resampler, and state MLP) outputs $\mathbf{z}_t$, which conditions the action generator to produce an action chunk.
    (b)~\textbf{Performance overview}. CoRP is competitive with much larger systems while using only 0.049B parameters.}
    \label{fig:architecture}
    \vspace{-10pt}
\end{wrapfigure}
 
\subsection{Policy Architecture}
\label{sec:architecture}
 
CoRP instantiates Eq.~\eqref{eq:decomposition} in two stages (Figure~\ref{fig:architecture}(a)): the representation extractor $f_\phi$ maps the multi-view observation $\mathbf{o}_t$, proprioceptive state $\mathbf{s}_t$, and instruction $\ell$ to a conditioning sequence $\mathbf{z}_t$, and the action generator $p_\theta$, a DiT-style Transformer trained with flow matching, models the distribution of $\mathbf{a}_{t:t+H-1}$ given $\mathbf{z}_t$.
 
\paragraph{Visual Encoder.}
Let $\mathbf{o}_t=\{\mathbf{o}_t^v\}_{v=1}^{V}$ denote the $V$ camera views.
Each view is encoded independently by a DINOv2-initialized ViT-S/14~\citep{oquab2024dinov2} with weights shared across views, yielding $P=256$ patch tokens at $224\times224$ resolution.
We discard the class token, pass the patch tokens to the resampler, and fine-tune the encoder jointly with the policy.
 
\paragraph{Language Encoder.}
We encode the instruction $\ell$ with a T5-Efficient-Tiny encoder~\citep{raffel2020t5,tay2022scale}, chosen because the instructions are short and drawn from a closed set, and fine-tune it jointly with the policy.
A LayerNorm and a linear adapter map its outputs to the policy width $d$, giving $\mathbf{u}\in\mathbb{R}^{M\times d}$, where $M$ is the padded instruction length.
A masked mean over the non-padding tokens gives the pooled embedding $\bar{\mathbf{u}}\in\mathbb{R}^{d}$, which conditions both the resampler and the action generator.
 
\paragraph{Vision--Language Fusion.}
Because the $P$ patch tokens per view are far more than the action generator needs, we compress them into $K\ll P$ tokens with a learned-query cross-attention resampler.
We linearly project the patch tokens to the policy width, giving $\mathbf{X}_t^v\in\mathbb{R}^{P\times d}$, and give each view its own trainable queries $\mathbf{Q}^v\in\mathbb{R}^{K\times d}$ ($K=48$ by default).
Before cross-attention, $\bar{\mathbf{u}}$ modulates the queries through a FiLM-style scale and shift~\citep{perez2018film}:
\begin{equation}
\widetilde{\mathbf{Q}}^v=\mathbf{Q}^v\odot\bigl(1+\boldsymbol{\gamma}(\bar{\mathbf{u}})\bigr)+\boldsymbol{\eta}(\bar{\mathbf{u}}),\qquad
\mathbf{r}_t^v=\operatorname{CrossAttn}\bigl(\widetilde{\mathbf{Q}}^v,\mathbf{X}_t^v\bigr),
\end{equation}
where $\operatorname{CrossAttn}$ is multi-head cross-attention (queries first, then keys and values) followed by LayerNorm, $\mathbf{r}_t^v\in\mathbb{R}^{K\times d}$ are the $K$ resampled tokens, and $\boldsymbol{\gamma},\boldsymbol{\eta}:\mathbb{R}^d\to\mathbb{R}^d$ are learned mappings whose outputs are broadcast across the $K$ queries.
 
A learned MLP $e_s$ maps the proprioceptive state to a single token $\widetilde{\mathbf{s}}_t=e_s(\mathbf{s}_t)\in\mathbb{R}^d$.
Concatenating the resampled visual tokens in view order, the language tokens, and the state token gives $\mathbf{z}_t\in\mathbb{R}^{(VK+M+1)\times d}$:
\begin{equation}
\mathbf{z}_t=f_\phi(\mathbf{o}_t,\mathbf{s}_t,\ell)
=\big[\,\mathbf{r}_t^{1};\ldots;\mathbf{r}_t^{V};\mathbf{u};\widetilde{\mathbf{s}}_t\,\big].
\label{eq:conditioning}
\end{equation}
The first $VK$ tokens form the \emph{visual prefix}, the only tokens of $\mathbf{z}_t$ that carry image information at the input to the action generator.
Keeping $K$ small therefore forces all visual information through a fixed-capacity bottleneck, which we test in Section~\ref{sec:exp-bottleneck}.
 
\paragraph{Action Generator.}
The action generator is an $L$-block, width-$d$ DiT-style Transformer ($L=1$ by default).
It linearly projects the noisy action chunk into $H$ action tokens with learned positional embeddings.
The interpolation time $\tau$ is embedded with sinusoidal encoding and an MLP, then added to every action token.
$\mathbf{z}_t$ is prepended to the action tokens as a prefix; the attention mask lets prefix tokens attend only to the prefix and action tokens attend to the prefix and all action tokens, keeping $\mathbf{z}_t$ independent of the noisy actions.
Each block also applies an adaptive LayerNorm (adaLN)~\citep{peebles2023scalable} with scale and shift computed from $\bar{\mathbf{u}}$, so the instruction conditions every layer rather than only the resampler.
The DiT then predicts a velocity for each action token (Section~\ref{sec:objective}).

\subsection{Training Objective}
\label{sec:objective}
 
\paragraph{Flow-Matching Loss.}
Let $(\mathbf{o}_t,\mathbf{s}_t,\ell,\mathbf{a})\sim\mathcal{D}$ be a training tuple from the demonstration dataset $\mathcal{D}$, where $\mathbf{a}\equiv\mathbf{a}_{t:t+H-1}\in\mathbb{R}^{H\times D_a}$ is the ground-truth action chunk and $D_a$ is the per-step action dimension.
We independently sample noise $\boldsymbol{\epsilon}\sim\mathcal{N}(\mathbf{0},\mathbf{I})$ of the same shape as $\mathbf{a}$ and an interpolation time $\tau\sim\operatorname{Beta}(1.0,1.5)$.
Here $\tau\in[0,1]$, distinct from the control step $t$, runs from noise ($\tau=0$) to data ($\tau=1$).
We form the point at time $\tau$ on the straight path and its constant target velocity:
\begin{equation}
\mathbf{a}^{\tau}=(1-\tau)\boldsymbol{\epsilon}+\tau\mathbf{a},\qquad
\mathbf{v}^\star=\mathbf{a}-\boldsymbol{\epsilon}.
\label{eq:interp}
\end{equation}
The action generator learns to predict this velocity, $\mathbf{v}_\theta$, by minimizing
\begin{equation}
\mathcal{L}_{\mathrm{fm}}=
\mathbb{E}\!\left[
\frac{1}{H D_a}
\left\|\mathbf{v}_\theta(\mathbf{a}^{\tau},\tau\mid\mathbf{z}_t)-\mathbf{v}^\star\right\|_F^2
\right],
\label{eq:fm_loss}
\end{equation}
where the expectation is over training tuples, $\boldsymbol{\epsilon}$, and $\tau$, and we exclude chunk positions padded beyond the end of an episode from both the norm and the normalization (Appendix~\ref{app:sim}).
 
At inference, we draw $\mathbf{a}^{\tau_0}\sim\mathcal{N}(\mathbf{0},\mathbf{I})$ and integrate with $N=10$ Euler steps on the uniform grid $\tau_n=n/N$:
\begin{equation}
\mathbf{a}^{\tau_{n+1}}=\mathbf{a}^{\tau_n}+\frac{1}{N}\,
\mathbf{v}_\theta(\mathbf{a}^{\tau_n},\tau_n\mid\mathbf{z}_t),
\qquad n=0,\ldots,N-1.
\label{eq:euler_sampling}
\end{equation}
The final sample $\mathbf{a}^{\tau_N}$ is the predicted chunk $\hat{\mathbf{a}}_{t:t+H-1}$, whose $H$ actions are all executed before replanning ($H=12$ on LIBERO and $H=50$ on RoboTwin~2.0; Appendix~\ref{app:sim}).
 
\paragraph{Task-Classification Loss.}
The flow-matching loss supervises only the actions, so it shapes the conditioning representation only indirectly, through gradients that reach $f_\phi$ via the action generator.
We therefore add an auxiliary task-classification loss that directly encourages the language-conditioned visual prefix to retain task-relevant information.
Let $c\in\{1,\ldots,C\}$ be the label of the current task and $\mathbf{h}^{(j)}\in\mathbb{R}^{VK\times d}$ the visual-prefix hidden states after the $j$-th DiT block, with $j=0$ denoting the input prefix.
For each depth $j$ in a set $\mathcal{J}$, a task head $g_\psi$ (an MLP shared across depths) applies LayerNorm to each token, averages over the $VK$ tokens, and produces logits
\begin{equation}
\boldsymbol{\xi}^{(j)}=g_\psi\!\left(\operatorname{mean}\bigl(\operatorname{LayerNorm}(\mathbf{h}^{(j)})\bigr)\right)\in\mathbb{R}^{C},
\label{eq:task_head}
\end{equation}
and $\mathcal{L}_{\mathrm{task}}$ averages the softmax cross-entropy $\operatorname{CE}$ over these depths:
\begin{equation}
\mathcal{L}_{\mathrm{task}}=\frac{1}{|\mathcal{J}|}\sum_{j\in\mathcal{J}}\operatorname{CE}\bigl(\boldsymbol{\xi}^{(j)},c\bigr).
\label{eq:task_loss}
\end{equation}
By default ($L=1$), $\mathcal{J}=\{1\}$.
The task head is used only during training: neither $c$ nor its logits condition action generation at inference.
 
\paragraph{Overall Objective.}
The overall training loss weights the two terms by $\lambda\geq0$ ($\lambda=0.01$ by default):
\begin{equation}
\mathcal{L}=\mathcal{L}_{\mathrm{fm}}+\lambda\mathcal{L}_{\mathrm{task}}.
\label{eq:total_loss}
\end{equation}
We minimize $\mathcal{L}$ jointly over $\phi$, $\theta$, and $\psi$ with a single AdamW optimizer~\citep{loshchilov2019decoupled} (Appendix~\ref{app:sim}), training all modules end-to-end, including the pretrained visual and language encoders (Section~\ref{sec:exp-adaptation} studies freezing the visual encoder).

\section{Experiments}
\label{sec:experiments}
Section~\ref{sec:exp-benchmark} shows CoRP is competitive with much larger systems, Sections~\ref{sec:exp-init}--\ref{sec:exp-bottleneck} isolate why by varying one property of $f_\phi$ at a time, and Section~\ref{sec:exp-real} evaluates CoRP on real robots.
 
\subsection{Setup}
\paragraph{Benchmarks.}
We evaluate on two simulation benchmarks, plus a precision peg-insertion task.

\emph{LIBERO}~\citep{liu2023libero} is a single-arm benchmark with four 10-task suites: Spatial, Object, Goal, and Long.
Its evaluation follows the demonstration distribution, so we use it for controlled comparisons (Sections~\ref{sec:exp-init}--\ref{sec:exp-bottleneck}), where differences reflect representation rather than robustness.

\emph{RoboTwin~2.0}~\citep{chen2025robotwin} is a 50-task bimanual benchmark (Aloha-AgileX) with Clean and Randomized settings; following \citet{bi2026motus}, we train on 2{,}500 clean and 25{,}000 randomized demonstrations, so Randomized tests in-distribution visual variation rather than zero-shot transfer.
We use it to test whether our findings extend to bimanual control.
 
\paragraph{Implementation.}
Unless stated otherwise, every variant uses the default configuration of Section~\ref{sec:method} ($K=48$, $L=1$, $d=768$, $\lambda=0.01$) and the same training recipe, changing only the factor under study.
Appendix~\ref{app:sim} lists the detailed training schedules and compute, using random seed 42 during training.

\begin{table}[h]
  \vspace{-8pt}
  \caption{\textbf{Success rates (\%) on LIBERO.}}
  \label{tab:libero_comparison}
  \centering
  \small
  \setlength{\tabcolsep}{3pt}
  \begin{tabular*}{\linewidth}{@{\extracolsep{\fill}}lcccccc}
    \toprule
    Method & Params (B) & Spatial & Object & Goal & Long & Avg. \\
    \midrule
    \multicolumn{7}{c}{\textit{With Vision-Language Model}} \\
    \midrule
    SmolVLA~\citep{shukor2025smolvla}
      & 2.25 & 93.0 & 94.0 & 91.0 & 77.0 & 88.8 \\
    $\pi_0$~\citep{black2025pi0}
      & 3.3 & 96.8 & \textbf{98.8} & 95.8 & 85.2 & 94.1 \\
    $\pi_{0.5}$~\citep{black2025pi05}
      & 3.3 & \textbf{98.8} & 98.2 & \textbf{98.0} & 92.4 & 96.9 \\
    OpenVLA~\citep{kim2024openvla}
      & 7.0 & 84.7 & 88.4 & 79.2 & 53.7 & 76.5 \\
    OpenVLA-OFT~\citep{kim2025openvlaoft}
      & 7.0 & 97.6 & 98.4 & 97.9 & \textbf{94.5} & \textbf{97.1} \\
    \midrule
    \multicolumn{7}{c}{\textit{With Video Generation Model}} \\
    \midrule
    Cosmos-Policy~\citep{kim2026cosmospolicy}
      & 2.0 & 98.1 & \textbf{100.0} & \textbf{98.2} & \textbf{97.6} & \textbf{98.5} \\
    Fast-WAM-Joint~\citep{yuan2026fastwam}
      & 6.0 & \textbf{99.6} & 99.4 & \textbf{98.2} & 96.8 & \textbf{98.5} \\
    Motus~\citep{bi2026motus}
      & 8.0 & 96.8 & 99.8 & 96.6 & \textbf{97.6} & 97.7 \\
    \midrule
    \multicolumn{7}{c}{\textit{Without Both}} \\
    \midrule
    ACT~\citep{zhao2023act}
      & 0.084 & 82.0 & 78.8 & 66.1 & 44.0 & 67.7 \\
    Diffusion Policy~\citep{chi2023diffusionpolicy}
      & 0.157 & 78.3 & 92.5 & 68.3 & 50.5 & 72.4 \\
    Octo~\citep{ghosh2024octo}
      & 0.093 & 78.9 & 85.7 & 84.6 & 51.1 & 75.1 \\
    ProgVLA~\citep{kim2026progvla}
      & 0.109 & 87.6 & 96.0 & 92.0 & 88.6 & 91.1 \\
    \midrule
    \textbf{Ours} ($L=1$, $\mathbf{d=180}$)
      & 0.037 & 97.6 & 99.4 & 93.2 & 85.4 & 93.9 \\
    \textbf{Ours} ($L=1$, $d=768$, default)
      & 0.049 & \textbf{98.2} & 99.4 & 95.8 & 94.4 & 97.0 \\
    \textbf{Ours} ($\mathbf{L=3}$, $d=768$)
      & 0.063 & \textbf{98.2} & \textbf{99.6} & \textbf{98.8} & \textbf{97.0} & \textbf{98.4} \\
    \bottomrule
  \end{tabular*}
\end{table}

\paragraph{Metrics.}
Our primary metric is task success rate (\%), following each benchmark's protocol~\citep{liu2023libero,chen2025robotwin}: 50 trials per task on LIBERO, reported per suite and averaged over 40 tasks; 100 trials per task on RoboTwin~2.0, averaged over 50 tasks per setting (per-task results in Table~\ref{tab:robotwin-full}).
We also report task-head accuracy (\%, Eq.~\eqref{eq:task_head}) on each evaluation episode's first frame, measuring how well the visual prefix identifies the task.
To measure action-relevant information without closed-loop execution, we freeze $f_\phi$ and train a lightweight probe to predict action chunks from the visual prefix~\citep{radosavovic2023robot,nikulin2025latent}, reporting its NMAE (per-dimension, normalized by the 1st--99th percentile range; lower is better) and $R^2$ (higher is better) on a held-out split (Eq.~\eqref{eq:probe_metrics}, details in~\ref{app:arch}).
\subsection{Can a Compact Policy Match Much Larger Systems?}
\label{sec:exp-benchmark}

\begin{wraptable}{r}{0.58\linewidth}
  \centering
  \vspace{-18pt}
  \caption{\textbf{Success rates (\%) on RoboTwin~2.0.}}
  \label{tab:robotwin_comparison}
  \small
  \setlength{\tabcolsep}{3pt}
  \begin{tabular}{lccc}
    \toprule
    Method & Params (B) & Clean & Rand. \\
    \midrule
    \multicolumn{4}{c}{\textit{With Vision-Language Model}} \\
    \midrule
    X-VLA~\citep{zheng2026xvla}                     & 0.9   & \textbf{72.9} & \textbf{72.8} \\
    $\pi_0$~\citep{black2025pi0}                    & 3.3   & 46.4 & 16.3 \\
    $\pi_{0.5}$~\citep{black2025pi05}               & 3.3   & 43.0 & 43.8 \\
    \midrule
    \multicolumn{4}{c}{\textit{With Video Generation Model}} \\
    \midrule
    Motus~\citep{bi2026motus}                       & 8.0   & \textbf{88.7} & \textbf{87.0} \\
    \quad w/o pretraining                           & 8.0   & 77.6 & 77.0 \\
    \begin{tabular}[c]{@{}l@{}}GigaWorld-\\~\citep{giga2026gigaworld}\end{tabular} & 11.4 & 86.4 & 85.0 \\
    \midrule
    \multicolumn{4}{c}{\textit{Without Both}} \\
    \midrule
    ACT~\citep{zhao2023act}                         & 0.084 & 29.7 & 1.7 \\
    Diffusion Policy~\citep{chi2023diffusionpolicy} & 0.157 & 28.0 & 0.6 \\
    \midrule
    \textbf{Ours}                                   & \textbf{0.049} & \textbf{75.8} & \textbf{73.4} \\
    \bottomrule
  \end{tabular}
\end{wraptable}
 
\textbf{The answer is YES on LIBERO, and partly on RoboTwin~2.0.}

Tables~\ref{tab:libero_comparison} and~\ref{tab:robotwin_comparison} compare CoRP with VLA, WAM, and compact policies that use neither prior; baseline results are taken from \citep{chen2025robotwin, giga2026gigaworld, bi2026motus, shukor2025smolvla, kim2024openvla, kim2025openvlaoft, yuan2026fastwam, kim2026progvla, kim2026cosmospolicy} (Appendix~\ref{app:baselines}).
On LIBERO, the default CoRP (0.049B) reaches 97.0\%, on par with $\pi_{0.5}$ (96.9\%, 3.3B) and OpenVLA-OFT (97.1\%, 7.0B), and 5.9 points above ProgVLA (91.1\%), the strongest prior-free compact policy.
Performance also increases with capacity, from 93.9\% ($d=180$, 0.037B) to 98.4\% ($L=3$; 0.063B), matching Fast-WAM-Joint and Cosmos-Policy (98.5\%); we nonetheless use the one-block default for ablations to keep them cheap and comparable.
 
On RoboTwin~2.0 (Clean/Randomized), CoRP (75.8\%/73.4\%) outperforms X-VLA (72.9\%/72.8\%) and $\pi_{0.5}$ (43.0\%/43.8\%) and approaches Motus without video pretraining (77.6\%/77.0\%) with about 160$\times$ fewer parameters, but stays below pretrained Motus (88.7\%/87.0\%) and GigaWorld-Policy (86.4\%/85.0\%).
Notably, Diffusion Policy and ACT, whose visual encoders are learned without a large-scale pretrained vision model, collapse in the Randomized setting (0.6\% and 1.7\%), whereas CoRP loses only 2.4 points, an early sign that the visual representation drives robustness.
 
\subsection{Does a Pretrained Visual Encoder Help?}
\label{sec:exp-init}
 
\textbf{The answer is YES: the DINOv2-initialized ViT-S/14 outperforms every other encoder variant by at least 14.9 points in average LIBERO success.}

To separate the effect of pretraining from that of architecture, we cross two initializations, pretrained or random, with two parameter-matched architectures in a $2\times2$ design.
The ViT-S/14~\citep{dosovitskiy2021image} (22.1M, default) is pretrained from a DINOv2 checkpoint~\citep{oquab2024dinov2} and the ResNet-34 (21.5M) from ImageNet-supervised weights~\citep{he2016deep,deng2009imagenet}; the ResNet's final feature map provides 49 spatial tokens per view to the resampler.
 
\begin{table}[h]
\vspace{-10pt}
  \caption{\textbf{Ablation study} on \textbf{visual encoder initialization and architecture}. }
  \label{tab:initialization_comparison}
  \centering
  \small
  \setlength{\tabcolsep}{5pt}
  \begin{tabular}{lccccccc}
    \toprule
    Method & Spatial (\%) & Object (\%) & Goal (\%) & Long (\%) & Average (\%) & NMAE & $R^2$ \\
    \midrule
    Random ResNet-34 & 91.2 & 97.8 & 72.8 & 66.6 & 82.1 & 0.12 & 0.54 \\
    Pretrained ResNet-34 & 85.4 & 98.8 & 37.0 & 76.8 & 74.5 & 0.09 & 0.73 \\
    Random ViT-S/14 & 84.4 & 95.6 & 86.0 & 46.4 & 78.1 & 0.11 & 0.59 \\
    \textbf{Pretrained ViT-S/14} & \textbf{98.2} & \textbf{99.4} & \textbf{95.8} & \textbf{94.4} & \textbf{97.0} & \textbf{0.07} & \textbf{0.81} \\
    \bottomrule
  \end{tabular}
\end{table}

Table~\ref{tab:initialization_comparison} shows that the pretrained ViT achieves the highest success rate on every suite (97.0\% on average) and the most action-decodable visual prefix (NMAE 0.07, $R^2=0.81$).
Neither factor alone explains this gain: with random initialization, the ViT is no better than the ResNet (78.1\% vs.\ 82.1\%), and ImageNet pretraining even lowers the ResNet's success (74.5\% vs.\ 82.1\%).
The gain therefore requires combining the two, consistent with DINOv2's patch-level pretraining providing fine-grained spatial features for the resampler to select from.
Separating this from the benefit of any pretrained ViT would require a supervised-pretrained ViT control, which we leave to future work.
All four variants identify the task with 100\% accuracy, yet the pretrained ResNet reaches only a 37.0\% success rate on LIBERO-Goal: these policies know which task to perform but lack the fine-grained visual information needed to do so.


\subsection{Is Pretraining Enough Without Adaptation?}
\label{sec:exp-adaptation}
 
\textbf{The answer is NO: freezing the pretrained encoder lowers average LIBERO success from 97.0\% to 77.2\%.}
If DINOv2 features were already sufficient for control, freezing them should cost little.
We test this by comparing two otherwise identical policies:

\begin{wrapfigure}{r}{0.48\textwidth}
    \centering
    \vspace{-10pt}
    \includegraphics[width=\linewidth]{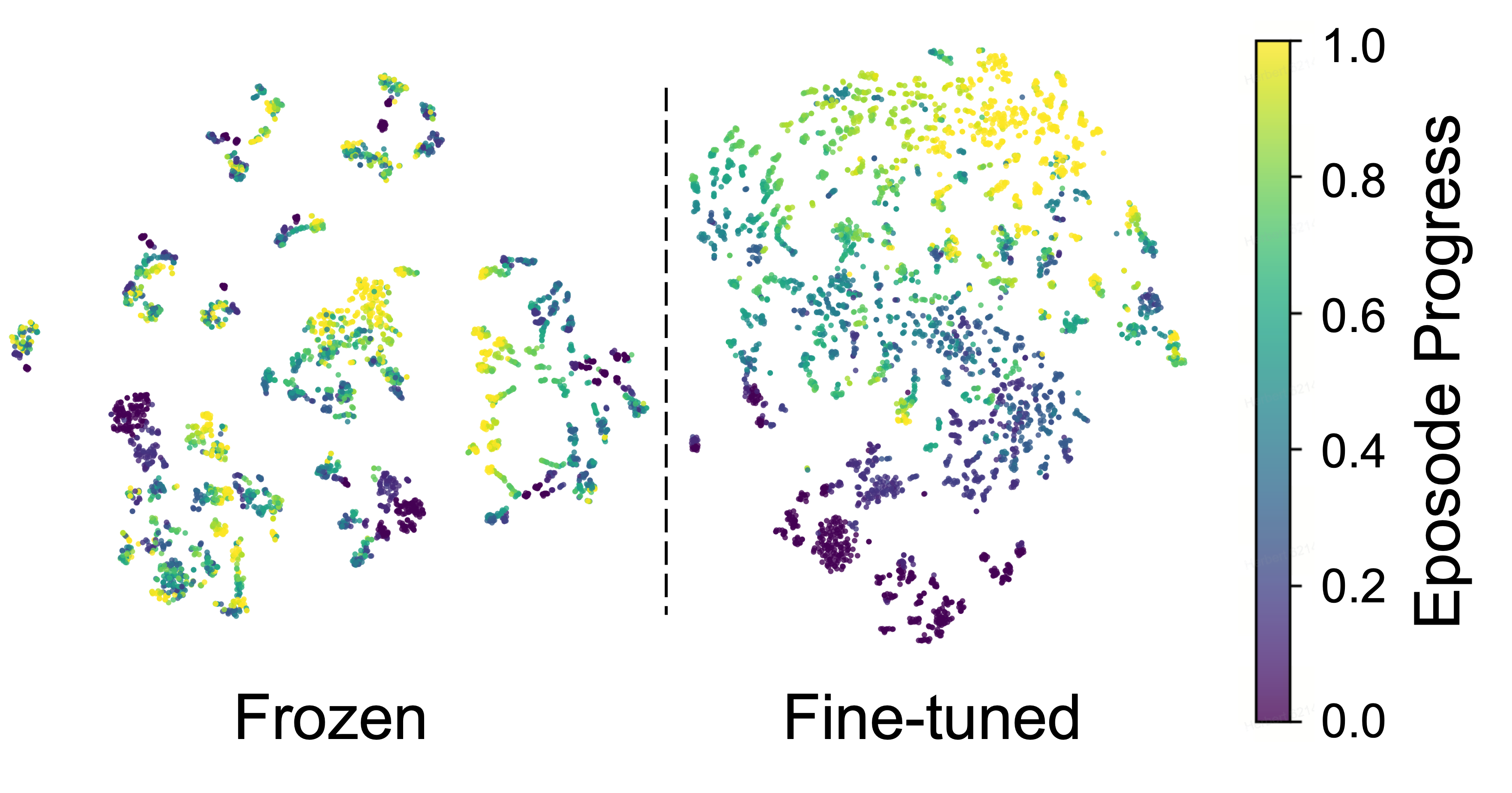}
    \vspace{-15pt}
    \caption{\textbf{t-SNE of visual-prefix features with and without fine-tuning.} Color indicates episode progress (purple: early, yellow: late). Fine-tuning groups features by stage and yields a clearer early-to-late progression.}
    \label{fig:finetune-tsne}
    \vspace{-15pt}
\end{wrapfigure}
\noindent\textbullet~\textbf{Frozen}: the DINOv2 encoder is fixed, and only the resampler, language adapter, and action generator are trained.

\noindent\textbullet~\textbf{Fine-tuned (default)}: all modules, including the encoder, are trained jointly.


The frozen policy (77.2\%) performs no better than a randomly initialized ViT trained end-to-end (78.1\%, Table~\ref{tab:initialization_comparison}), so pretraining pays off only when the features adapt to the task. The probe shows why: freezing reduces $R^2$ from 0.81 to 0.24 and doubles NMAE (0.07 to 0.16), indicating that frozen features retain much less action-relevant detail, and t-SNE analysis~\citep{vandermaaten2008visualizing} (Figure~\ref{fig:finetune-tsne}) shows that, after fine-tuning, representations at similar episode stages cluster more closely. They are also more clearly ordered from early to late stages.

\begin{wraptable}{r}{0.38\textwidth}
  \vspace{-20pt}
  \caption{\textbf{Ablation study} on whether to
  \textbf{fine-tune the visual encoder}.}
  \label{tab:frozen_finetune_comparison}
  \centering
  \small
  \setlength{\tabcolsep}{3pt}

  \begin{tabular}{lccc}
    \toprule
    Method & Average (\%) & NMAE & $R^2$ \\
    \midrule
    Frozen & 77.2 & 0.16 & 0.24 \\
    \textbf{Fine-tuned} & \textbf{97.0}
      & \textbf{0.07} & \textbf{0.81} \\
    \bottomrule
  \end{tabular}

  \vspace{-10pt}
\end{wraptable}

\subsection{Do We Need Language Conditioning?}
\label{sec:exp-conditioning}
 
\textbf{The answer is: it depends. Language is needed only when the observation does not uniquely specify the goal $\kappa$, i.e., which task is consistent with the current scene.}

We compare the default policy with one that removes the language encoder and tokens $\mathbf{u}$, with FiLM and adaLN receiving a learned constant conditioning in place of $\bar{\mathbf{u}}$.

\begin{wraptable}{r}{0.75\textwidth}
\vspace{-15pt}
  \caption{\textbf{Language-conditioning ablation on LIBERO} (\%).}
  \label{tab:lang_cond_libero_comparison}
  \centering
  \small
  \setlength{\tabcolsep}{3pt}
  \begin{tabular}{@{}l*{5}{cc}@{}}
    \toprule
    & \multicolumn{2}{c}{Spatial}
    & \multicolumn{2}{c}{Object}
    & \multicolumn{2}{c}{Goal}
    & \multicolumn{2}{c}{Long}
    & \multicolumn{2}{c}{Average} \\
    \cmidrule(lr){2-3}\cmidrule(lr){4-5}\cmidrule(lr){6-7}
    \cmidrule(lr){8-9}\cmidrule(l){10-11}
    Method
    & Acc. & S.R. & Acc. & S.R. & Acc. & S.R. & Acc. & S.R. & Acc. & S.R. \\
    \midrule
    Without Lang. & 90.4 & 87.0 & \textbf{100.0} & \textbf{100.0} & 8.6 & 9.2
                  & 89.0 & 83.4 & 72.0 & 69.9 \\
    \textbf{With Lang.} & \textbf{100.0} & \textbf{98.2} & \textbf{100.0} & 99.4
               & \textbf{100.0} & \textbf{95.8} & \textbf{100.0}
               & \textbf{94.4} & \textbf{100.0} & \textbf{97.0} \\
    \bottomrule
  \end{tabular}
  \vspace{-5pt}
\end{wraptable}

Removing language is catastrophic only on LIBERO-Goal, where success drops from 95.8\% to 9.2\%, while the other suites lose at most 11.2 points (Table~\ref{tab:lang_cond_libero_comparison}).
Without the instruction, the policy can only model the goal-marginal distribution $p(\mathbf a_t\mid\mathbf o_t,\mathbf s_t) =\sum_\kappa p(\mathbf a_t\mid\mathbf o_t,\mathbf s_t,\kappa) p(\kappa\mid\mathbf o_t,\mathbf s_t)$; when the scene is shared across goals, as in LIBERO-Goal, $p(\kappa\mid\mathbf{o}_t,\mathbf{s}_t)$ is nearly uniform, and the policy effectively commits to a random goal.

\begin{wraptable}{r}{0.4\textwidth}
\vspace{-15pt}
  \caption{\textbf{Language-conditioning ablation on RoboTwin~2.0} (\%).}
  \label{tab:lang_cond_robotwin_comparison}
  \centering
  \footnotesize
  \setlength{\tabcolsep}{2.5pt}
  \begin{tabular}{@{}l*{5}{cc}@{}}
    \toprule
    & \multicolumn{2}{c}{Clean}
    & \multicolumn{2}{c}{Randomized} \\
    \cmidrule(lr){2-3}\cmidrule(lr){4-5}
    Method
    & Acc. & S.R. & Acc. & S.R. \\
    \midrule
    \textbf{Without Lang.} & \textbf{99.96} & \textbf{78.50} & \textbf{99.92} & \textbf{76.06} \\
    With Lang. & 99.78 & 75.78 & 99.58  & 73.36\\
    \bottomrule
  \end{tabular}
  \vspace{-10pt}
\end{wraptable}

The no-language policy's task head confirms this: its first-frame accuracy closely tracks its success rate on every suite (Table~\ref{tab:lang_cond_libero_comparison}), and on Goal both are near the 10\% chance level for a 10-task suite (8.6\% and 9.2\%).
With language, the head reaches 100.0\% on every suite, since the instruction, injected into the visual prefix through FiLM, resolves the goal the scene leaves ambiguous.
The t-SNE analysis (Appendix Figure~\ref{fig:lang-conditioning-tsne-full}) illustrates that, without language conditioning, the Goal tasks collapse into a single cluster, whereas other tasks remain separable.

Conversely, on RoboTwin~2.0, where each scene identifies its task, removing language raises success by about 2.7 points in both settings, with task accuracy near 100\% either way (Table~\ref{tab:lang_cond_robotwin_comparison}); here the instruction adds no information.
Language thus acts as a disambiguator rather than a general source of task knowledge: it is essential when the scene leaves the goal ambiguous, and unnecessary or even slightly harmful when the scene determines it.

\subsection{Does the Information Bottleneck Help?}
\label{sec:exp-bottleneck}

 
\textbf{The answer is yes, but not every bottleneck formulation helps}: a hard, instruction-aware token budget raises average LIBERO success from 83.2\% to 97.0\%, whereas a variational bottleneck lowers success with or without it.
The information bottleneck (IB) principle~\citep{tishby1999information} seeks a representation $Z$ of the input $X$ that is maximally informative about a target $Y$ while discarding the rest of $X$, i.e., $\max_{p(Z\mid X)} I(Z;Y)-\beta_{\mathrm{IB}}I(Z;X)$ with $\beta_{\mathrm{IB}}>0$; here, $X$ is the visual input, $Z$ the visual prefix, and $Y$ the action chunk.
A common tractable surrogate is the deep variational information bottleneck (VIB)~\citep{alemi2017deep}, which adds $\beta_{\mathrm{IB}}D_{\mathrm{KL}}(q(Z\mid X)\,\Vert\,\mathcal{N}(\mathbf{0},\mathbf{I}))$ to a prediction loss, where $q(Z\mid X)$ is a Gaussian encoder.
In our VIB variants, we add this KL term on each visual token to our training loss (Eq.~\eqref{eq:total_loss}), with $\beta_{\mathrm{IB}}=4\times10^{-8}$ (Appendix~\ref{app:resampler}).
However, the KL penalty is task-agnostic: it compresses every token toward the same prior, regardless of which visual details the current instruction requires.
Our resampler instead imposes a hard, instruction-aware bottleneck: capacity is limited to $K\ll P$ tokens but allocated according to the instruction through the FiLM-modulated queries.
We cross the two choices in a $2\times2$ design: with or without the resampler, and with or without VIB.
Without the resampler (\emph{All Tokens}), all $P=256$ projected patch tokens per view enter the visual prefix.
In this case, $\bar{\mathbf{u}}$ directly modulates every patch token.
 

\begin{table}[h]
  \vspace{-10pt}
  \caption{\textbf{Ablation study on information bottleneck.} {\footnotesize (Acc.: task-head accuracy; S.R.: success rate.)}}
  \label{tab:ib_comparison}
  \centering
  \small
  \setlength{\tabcolsep}{3pt}
  \begin{tabular*}{\linewidth}{@{\extracolsep{\fill}}l*{5}{cc}@{}}
    \toprule
    & \multicolumn{2}{c}{Spatial}
    & \multicolumn{2}{c}{Object}
    & \multicolumn{2}{c}{Goal}
    & \multicolumn{2}{c}{Long}
    & \multicolumn{2}{c}{Average} \\
    \cmidrule(lr){2-3}\cmidrule(lr){4-5}\cmidrule(lr){6-7}
    \cmidrule(lr){8-9}\cmidrule(l){10-11}
    Method
    & Acc. & S.R. & Acc. & S.R. & Acc. & S.R. & Acc. & S.R. & Acc. & S.R. \\
    \midrule
    All Tokens           & 93.8 & 89.6 & 100.0 & \textbf{99.6} & 91.4 & 55.8 & 100.0 & 87.6 & 96.3 & 83.2 \\
    All Tokens + VIB     & 99.8 & 87.8 & 100.0 & 99.4 & 99.9 & 33.6 & 100.0 & 90.2 & 99.9 & 77.8 \\
    \textbf{Partial Tokens} & \textbf{100.0} & \textbf{98.2} & \textbf{100.0} & 99.4 & \textbf{100.0} & \textbf{95.8} & \textbf{100.0} & \textbf{94.4} & \textbf{100.0} & \textbf{97.0} \\
    Partial Tokens + VIB & 91.6 & 87.8 & 99.9 & 99.2 & 69.6 & 33.0 & 95.5 & 83.6 & 89.2 & 75.9 \\
    \bottomrule
  \end{tabular*}
\end{table}
 
The gain over All Tokens is concentrated on LIBERO-Goal (55.8\% to 95.8\%, Table~\ref{tab:ib_comparison}), the suite where the instruction alone specifies the goal.
To quantify how strongly the visual prefix depends on the instruction, we keep the observation fixed, swap the instruction for another task's, and measure the prefix's relative change ($\Delta_{prefix}$, Appendix~\ref{app:resampler}): 49.7\% for Resampled but 5.4\% for All Tokens.
Thus, the resampler selects instruction-specific visual information, whereas passing all tokens leaves the visual prefix nearly instruction-invariant and forces the action generator to resolve the goal on its own.

\begin{wrapfigure}{r}{0.46\textwidth}
    \centering
    \vspace{-10pt}
    \includegraphics[width=\linewidth]{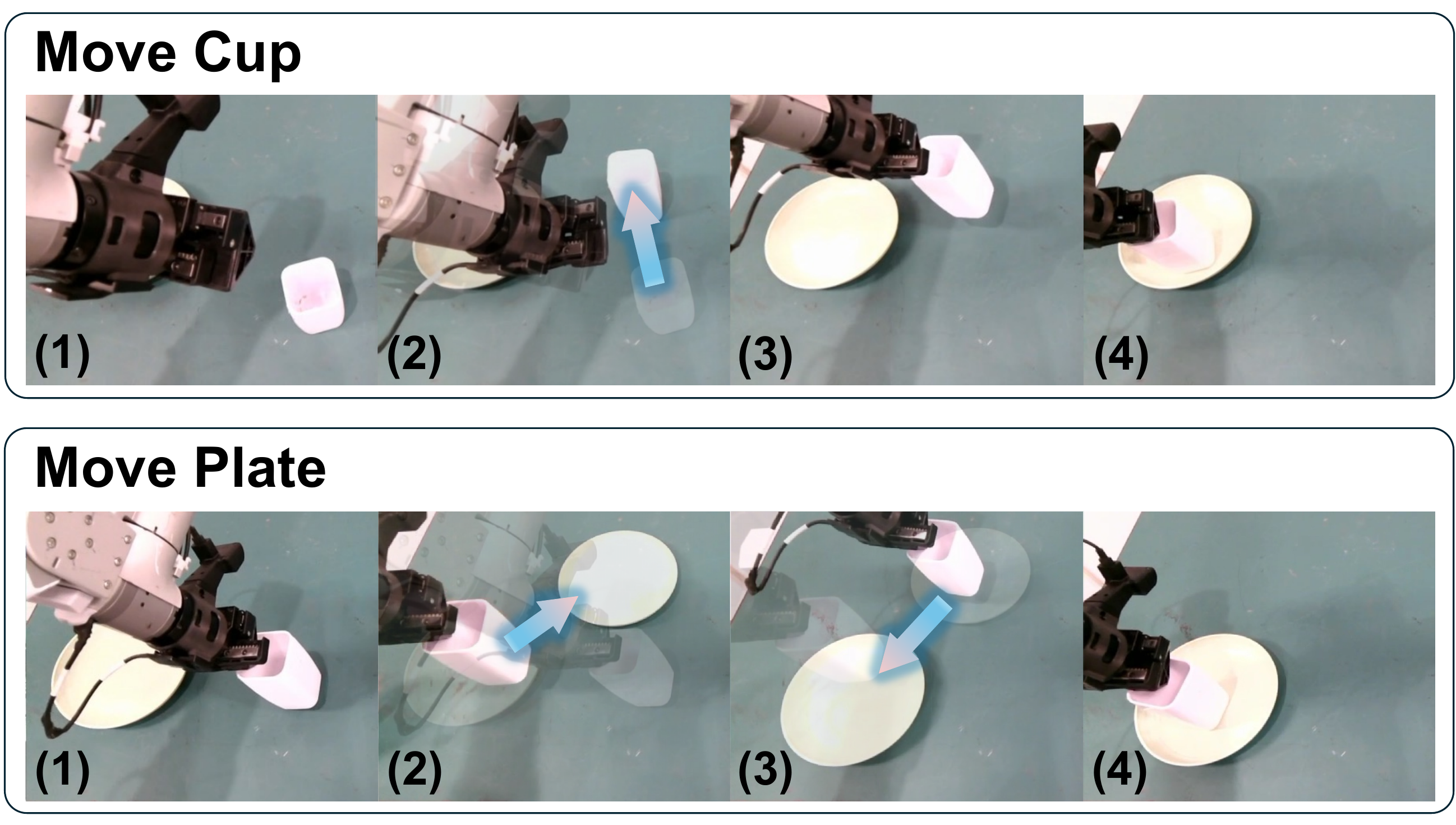}
    \vspace{-15pt}
    \caption{\textbf{Dynamic adjustment in the pick-cup task.}
            The robot adapts to cup and plate displacements during execution.}
    \label{fig:dynamic_adjustment}
    \vspace{-20pt}
\end{wrapfigure}

Adding VIB undoes this selection: on the resampled prefix, it reduces the instruction-induced change from 49.7\% to 13.6\% and Goal success from 95.8\% to 33.0\%; on All Tokens, it lowers average success from 83.2\% to 77.8\%.
Notably, All Tokens + VIB still identifies the Goal task with 99.9\% accuracy yet succeeds only 33.6\% of the time: the KL penalty preserves which task to perform but suppresses the instruction-specific visual detail needed to perform it.
A tight, instruction-aware budget therefore outperforms a soft, task-agnostic one.

\subsection{Can CoRP Work on Real Robots?}
\label{sec:exp-real}

Finally, we test whether CoRP transfers to physical hardware on three bimanual tasks that probe closed-loop adaptation, arm coordination, and precision.
Unlike in simulation, we pretrain the unchanged 48.9M-parameter CoRP on more than 10{,}000 hours of real-world data covering 5{,}887 tasks on our dual-arm robots, then fine-tune it on 10 tasks (Appendix~\ref{app:real}).
We evaluate three of these tasks over 10 trials each with randomized initial object placements.
The policy uses three cameras, one on the head and one on each arm, and runs synchronously at about 7\,ms per chunk.
 

\textbf{Pick cup.}
The robot grasps a cup by its thin wall and places it on a plate, both at random positions on the table.
CoRP succeeds in 12/12 trials. To test closed-loop adaptation, we manually move the cup

\begin{wrapfigure}{r}{0.67\textwidth}
    \centering
    \vspace{-0pt}
    \includegraphics[width=\linewidth]{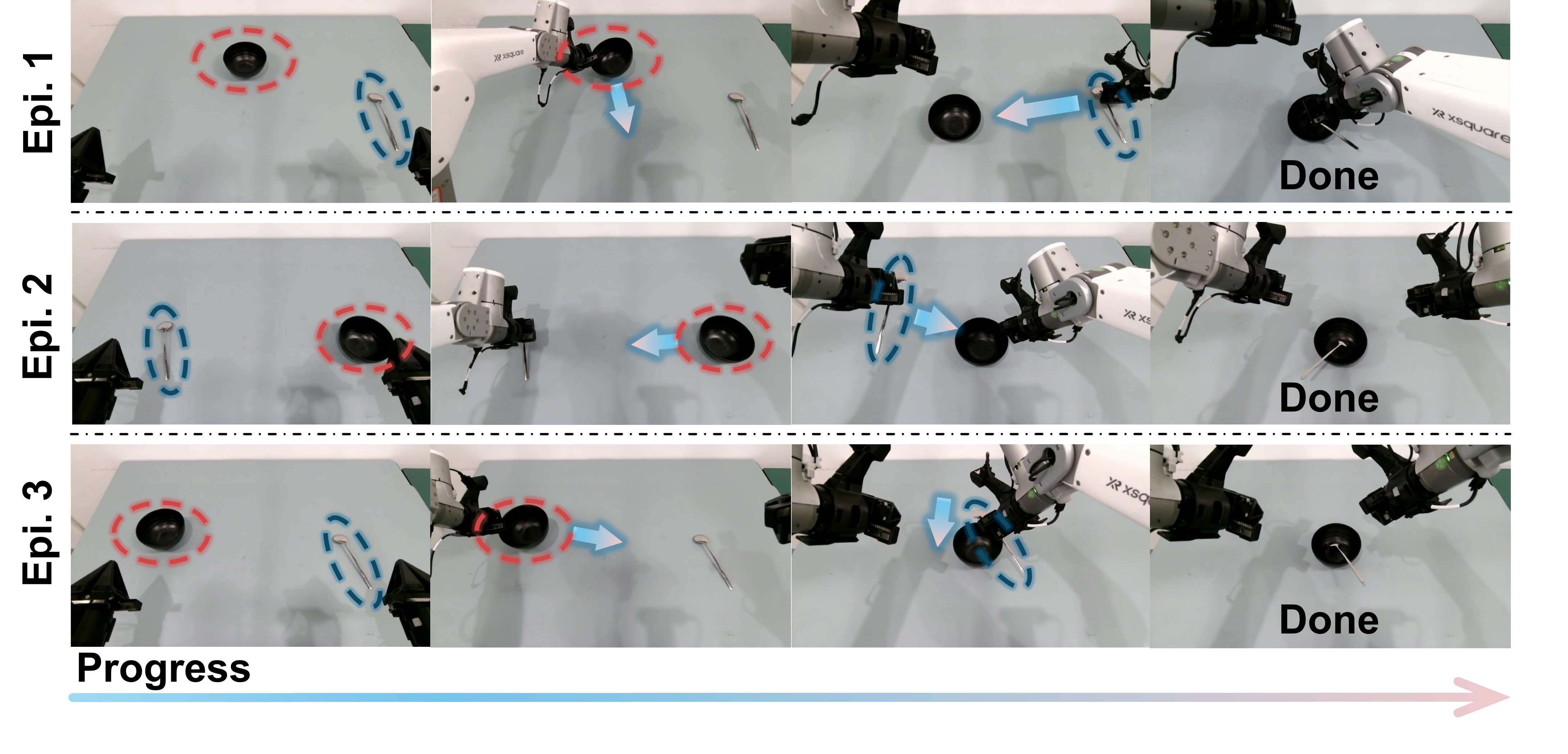}
    \vspace{-20pt}
    \caption{\textbf{Bimanual coordination in the place-spoon task}, shown from three initial configurations.}
    \label{fig:place_spoon}
    \vspace{-15pt}
\end{wrapfigure}
\noindent or plate during execution in 8 trials.
As Figure~\ref{fig:dynamic_adjustment} shows, the policy re-targets the displaced object within the next action chunk and completes the task in 8/8 perturbed trials.
This shows that CoRP reacts to new observations rather than replaying a memorized trajectory.

\textbf{Place spoon.}
The robot picks up a spoon from the table and places it in a bowl.
Because the bowl may start outside the spoon-holding arm's reach, the other arm must first move it into reach.
Across 10 initial configurations (three shown in Figure~\ref{fig:place_spoon}), CoRP succeeds 9/10.

\begin{wrapfigure}{r}{0.45\textwidth}
    \centering
    \vspace{-10pt}
    \includegraphics[width=\linewidth]{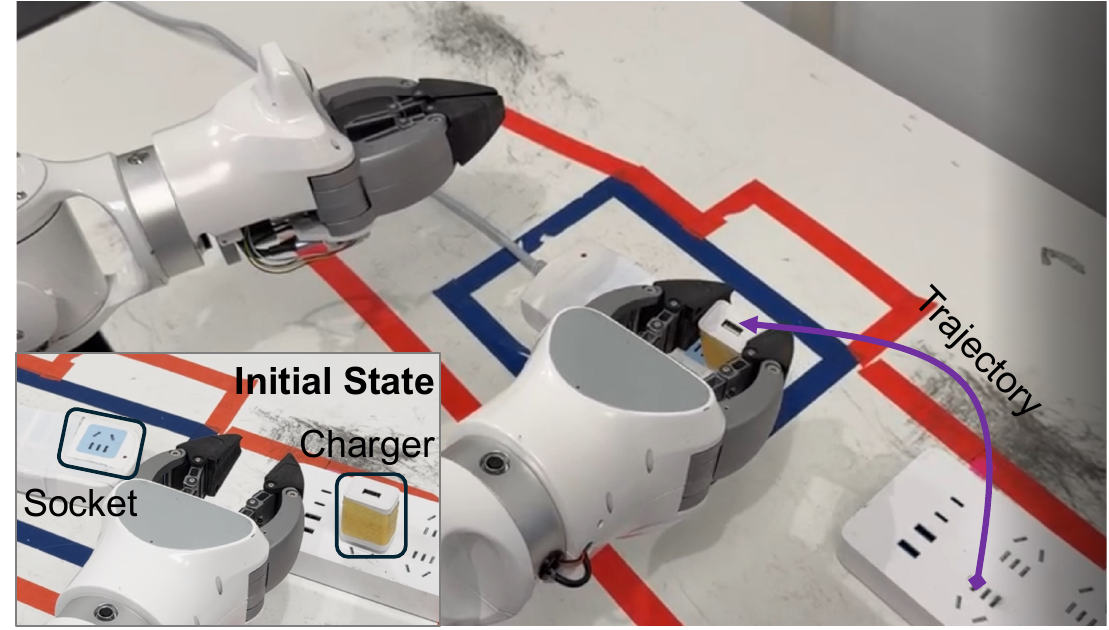}
    \vspace{-15pt}
    \caption{\textbf{Precision manipulation in the plug-insertion task.} The robot unplugs a charger from the right socket and inserts it into the left socket.}
    \label{fig:plug_insert}
    \vspace{-5pt}
\end{wrapfigure}
\textbf{Plug insertion.}
The robot unplugs a charger from the right socket, transfers it, and inserts it into the left socket (Figure~\ref{fig:plug_insert}).
Successful insertion requires aligning the pins with the socket openings in both position and orientation and maintaining that alignment as the pins enter.
It thus tests precision manipulation under geometric constraints, complementing the simulated peg-insertion task in Appendix~\ref{app:sim} (8.0\, mm peg, 8.1\, mm hole), where CoRP reaches 84\% success.
CoRP completes the insertion in 2 trials. Since the architecture is unchanged from simulation, these results suggest that the same compact, representation-centered design scales to real-world pretraining and deployment.

\section{Conclusion and Limitations}
\label{sec:conclusion}
We presented CoRP, a 48.9M-parameter manipulation policy without a VLM or video-generative prior, and used it as a controlled testbed to ask where a multi-task policy's performance comes from.
By holding the action generator fixed and varying one property of the representation extractor at a time, we find that a compact policy works when its visual representation is pretrained, task-adapted, and compressed: removing any one of these properties costs 14--20 points on LIBERO, while language matters only when the scene leaves the goal ambiguous.
With these properties, CoRP is competitive with systems up to 163$\times$ larger on LIBERO and RoboTwin~2.0 and transfers unchanged to real bimanual robots.
Across our ablations, the weaker variants still identify the task but fail to perform it, suggesting that compact policies lack not task knowledge but fine-grained, instruction-specific visual detail.
Our study has several limitations that are worth exploring in future work.
 
\textbf{Closed-set evaluation.}
Our benchmarks measure in-distribution control on closed task sets; we do not evaluate the open-world semantic and physical generalization that VLAs and WAMs target.
Whether pretrained, task-adapted, and compressed representations remain sufficient in that regime, or must be combined with such priors, is an open question.
 
\textbf{Attribution of the pretraining benefit.}
Our encoder ablation lacks a supervised-pretrained ViT, so it cannot separate the benefit of DINOv2's patch-level pretraining from that of any pretrained ViT; adding this control is the next direct step.
 
\textbf{Real-world evidence.}
Our real-robot experiments cover three tasks, include no real-robot baselines, and follow large-scale pretraining; they show that CoRP transfers to hardware but do not isolate the role of the representation as our simulation ablations do.
Repeating the controlled interventions on real hardware would close this gap.

\section*{AI USE STATEMENT}
In compliance with the ICLR 2027 policy, we acknowledge using large language models (LLMs) as assistive tools in preparing this paper.
We used LLMs to polish the writing and help develop the software, including debugging and generating scripts for data preprocessing and visualization.
The authors carefully reviewed and edited all final text to ensure it accurately reflects our research and contributors. 
The authors also developed the core algorithmic implementation of our proposed method.
Overall, the LLMs served mainly as a productivity and quality-enhancement tool. 

\section*{Reproducibility Statement}
To support reproducibility, Section~\ref{sec:method} describes CoRP's architecture and training objectives, while Section.~\ref{sec:experiments} specifies the benchmarks and evaluation metrics.
Sections~\ref{sec:exp-init}--\ref{sec:exp-bottleneck} present controlled ablations on visual encoder initialization, fine-tuning, language conditioning, and representation compression.
Our simulation experiments use the publicly available LIBERO and RoboTwin~2.0 datasets.
The real-world training datasets are proprietary and cannot be released due to commercial restrictions, which limits independent reproduction of the real-world results.
We will release the source code and configuration files for the simulation experiments with the camera-ready version to facilitate their reproduction.

\bibliography{references}
\bibliographystyle{iclr2027_conference}

\clearpage
\appendix

\section{Additional Experimental Details}
We provide additional experimental details for all experiments in this paper.

\subsection{Policy Architecture}\label{app:arch}

\paragraph{Visual encoder.} The visual backbone loaded from DINOv2 ViT-S/14. 
At 224 $\times$ 224, the 14-pixel patch size yields a 16 $\times$ 16 = 256 patch-token grid per camera, each 384 wide. 
The class token and any register tokens are not passed to the policy. 
The same backbone weights are shared across all views; the resampler distinguishes views using learned view embeddings. 
The default policy fine-tunes the backbone. 
The image encoder is not a video encoder and receives no frame history in this paper.

\paragraph{Language encoder.}
We load the language backbone from Google T5-Efficient-Tiny.
We only load \textit{transformers.T5TokenizerFast} and \textit{transformers.T5EncoderModel} and drop the decoder part.
Tokenization uses padding to a fixed maximum length of 32 and truncation. 
The attention mask is used for the masked mean, and the projected padded token rows are set to zero. 
The encoder has a model width of 256, a feed-forward width of 1024, four encoder layers, four attention heads, and $d_{kv}=64$. A LayerNorm and a $256 \rightarrow 768$ linear adapter produce the policy-width language tokens.

\paragraph{Resampler and FiLM.}
\label{app:resampler}
Each view is projected from 384 to 768 and compressed to 48 learned query tokens. 
The queries use eight attention heads in one MultiheadAttention layer, followed by LayerNorm. 
This resampler implementation has no separate FFN or dropout and no explicit resampler residual branch. 
Query parameters are separate for each view unless a view-role alias is supplied. 
The number of visual prefixes equals the number of views times 48 learned query tokens.

To measure the resampled visual prefix's sensitivity to the language instruction, we introduce $\Delta_{prefix}$, which we use in Section~\ref{sec:exp-bottleneck}.
Let $\mathbf{R_i} = \big[\,{\mathbf{r}_t^{1};\ldots;\mathbf{r}_t^{V}}\big]=R(\mathbf{o}_i, \mathbf{s}_i, \ell_i)$ denote a visual prefix produced by the resampler using the original instruction $\ell_i$ and  $\tilde{\mathbf{R}}_i=R(\mathbf{o}_i, \mathbf{s}_i, \tilde{\ell}_i)$ denote the visual prefix after replacing the instruction with one from a different instruction in dataset.
The relative change is defined as:
\begin{equation}
    \Delta_{prefix} = \sqrt{\frac{\sum_{i=1}^N||\tilde{\mathbf{R}}_i - \mathbf{R}_i||_F^2}{\sum_{i=1}^N||\mathbf{R}_i||_F^2}},
\end{equation}
where $N=8000$.
For fairness, we use the same instruction-swap pairs for all variants in the experiment. 

\paragraph{Action generator.}
We use a $D_a \rightarrow 768$ projection to transform noisy actions into tokens.
We follow the widely used sinusoidal time embedding to inject interpolation time.
Let $P$ be the complete conditioning prefix (visual tokens, language tokens, and the state token) and let $A$ be the noisy action suffix. 
Prefix query rows can attend to every valid prefix key but cannot attend to any noisy action key. 
Every action query row can attend to every valid prefix key and to every action key, including the other action tokens. 
Thus, action tokens are bidirectional, not causal. 
The language mask removes padding from the key side.
The task head pools only the visual-prefix positions: at layer 0 this is direct visual information.
After a transformer block, the prefix attention rule allows those visual positions to carry information from the language and state tokens as well.


\paragraph{Variational information bottleneck.}
The VIB is an archived ablation, not a module in the current policy.
We apply VIB after the visual resampler, before concatenating language tokens.
For each visual token $\mathbf{x}$, 
\begin{equation}
\mathbf{\mu}=W_\mu \mathbf x+b_\mu, \quad \log \sigma^2 = W_\sigma \mathbf x  + b_\sigma, \quad \mathbf z = \mathbf \mu + \epsilon \odot \exp(\frac{1}{2}\log\sigma^2).
\end{equation}
We adopt a standard normal Gaussian as the prior, and $\beta_{IB} = 4\times 10^{-8}$ in experiments.

\paragraph{Action probe.}
To quantify whether a representation is useful for action prediction, we freeze each model's representation extractor $f_\phi$ and train an action probe, following prior robot representation studies~\citep{radosavovic2023robot,nikulin2025latent}.
The probe is trained mainly to evaluate representation quality rather than to control robots.
Given a visual prefix, the probe predicts the robot's next action chunk. 
We make a separate train/test split of the LIBERO demonstrations.
For each task, we train the probe on 80\% of the episodes and test it on the remaining 20\%.
This tests whether the frozen representation contains action information that can be read out on episodes in the test set.
More accurate predictions indicate that the representation preserves higher-quality information for manipulation. 
We report NMAE and $R^2$ as the metrics:
\begin{equation}
NMAE=mean 
\begin{pmatrix}\frac{\mathbf{|\hat a} - \mathbf a|}{\mathbf q_{99} - \mathbf q_{01}}
\end{pmatrix}, 
\quad 
R^2=1-\frac{\|\hat{\mathbf a}-\mathbf a\|_2^2} {\|\mathbf a-\bar{\mathbf a}\|_2^2}, 
\label{eq:probe_metrics}
\end{equation}
where $\mathbf a$ is the ground-truth action chunk, $\hat{\mathbf a}$ is the probe prediction, $\bar{\mathbf a}$ is the mean action chunk, and $\mathbf q_{01}$ and $\mathbf q_{99}$ are the per-dimension action percentiles. 
The action probe's training target approximates the full $H\times D_a$ action chunk. 

\subsection{Inference Cost}\label{app:infer}
\paragraph{Time cost.}
We tested the inference time on an Nvidia 5090 GPU.
The full inference time cost for the default configuration without any optimization is $7.01\,\mathrm{ms}$, where the representation extractor $f_\phi$ costs about $3.43\,\mathrm{ms}$, while the action generator costs about $3.08\,\mathrm{ms}$.
We also tried to accelerate inference using CUDA Graph capture and replay to reduce CPU-side kernel-launch overhead.
After acceleration, we reach $5.79\,\mathrm{ms}$ for inference.
All results in this paper are inferred by the policy without acceleration, since it is already fast enough.

\paragraph{Memory cost.}
The policy's model weights occupy a total of $95.4\,\mathrm{MB}$ of GPU memory, with DINOv2-S, T5, and DiT accounting for $42.1\,\mathrm{MB}$, $23.7\,\mathrm{MB}$, and $23.7\,\mathrm{MB}$, respectively, and the remaining components contributing $5.9\,\mathrm{MB}$.
All model parameters are stored in BF16 precision.

\subsection{Simulation}
\label{app:sim}

\paragraph{Implementation Details.}
We resize all images to $224\times224$ in both training and evaluation. 
Our visual encoder is DINOv2 ViT-S/14~\citep{oquab2024dinov2} (22.1M parameters), and we fine-tune it by default. 
A cross-attention resampler maps its patch features to 48 tokens per camera, each with width 768. 
We encode instructions with the Google T5-Efficient-Tiny language encoder~\citep{raffel2020t5,tay2022scale}, which has 11.4M parameters and is fine-tuned with the policy.
A trainable LayerNorm and linear layer map its output to width 768.
Both the visual and language encoders are the smallest in their respective families.

The default action generator has one DiT-style block~\citep{peebles2023scalable} with width 768 and about 12.2M parameters.
A small task head maps the mean visual token to discrete task types.
We train it with cross-entropy weight $\lambda=0.01$. 
The complete policy has about 48.9M parameters.

We train for 80k steps on LIBERO and 120k steps on RoboTwin~2.0.
We train directly on the original LIBERO dataset, while following Motus's~\citep{bi2026motus} recipe to use 2,500 demonstrations collected in clean scenes (50 per task) plus 25,000 demonstrations collected in randomized scenes (500 per task).
Both settings use AdamW~\citep{loshchilov2019decoupled} with a learning rate of $5\times10^{-5}$, batch size 392, BF16 training, gradient clipping at 1.0, and an EMA decay of 0.999.
For flow-matching training~\citep{lipman2023flowmatching}, we sample times from $\operatorname{Beta}(1.0,1.5)$.
We use $2\times$ A800 for training; LIBERO costs about 8 hours, while RoboTwin~2.0 costs about 13 hours.
At test time, we generate each action chunk with 10 Euler steps.

\begin{figure}[h]
    \centering
    \includegraphics[width=1.0\textwidth]{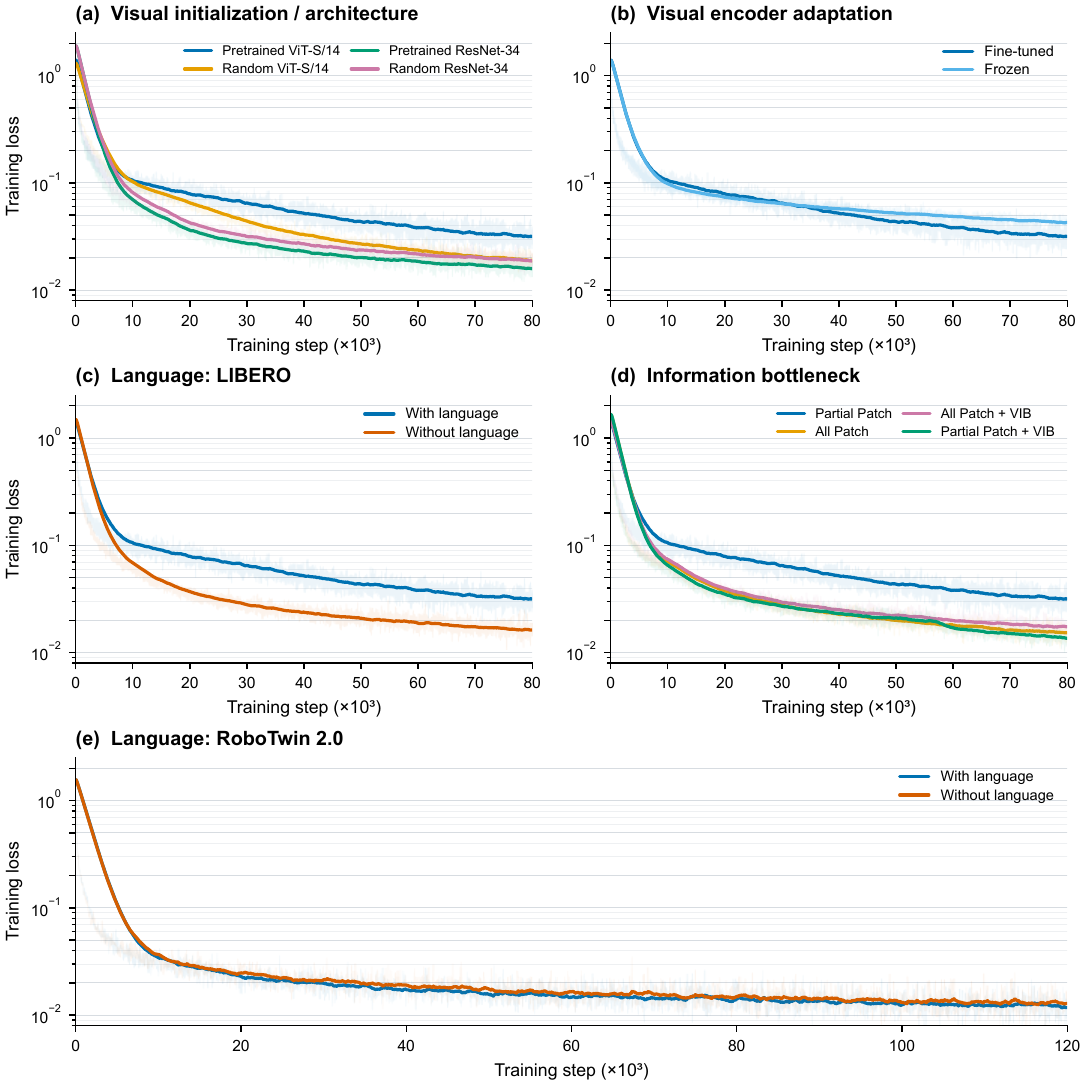}
    \caption{
    \textbf{Training loss does not predict policy performance.}
    Training loss (log scale) for the ablations in Sections~\ref{sec:exp-init} (a), \ref{sec:exp-adaptation} (b), \ref{sec:exp-conditioning} (c: LIBERO, e: RoboTwin~2.0), and \ref{sec:exp-bottleneck} (d).
    In (a), (c), and (d), the best-performing configuration (dark blue) reaches the highest loss; freezing the encoder (b) is the exception, raising loss and lowering success.
    With and without language, the RoboTwin~2.0 curves (e) nearly coincide.
    }
    \label{fig:ablation_training_losses}
\end{figure}

Figure~\ref{fig:ablation_training_losses} compares the training loss curves across the ablation studies.
Lower training loss does not necessarily translate into better policy performance.
Notably, the best-performing configurations often exhibit the highest training losses.
We do not report validation losses because we train on the full dataset without a held-out validation split.

\begin{figure}[h]
    \centering
    \includegraphics[width=1.0\textwidth]{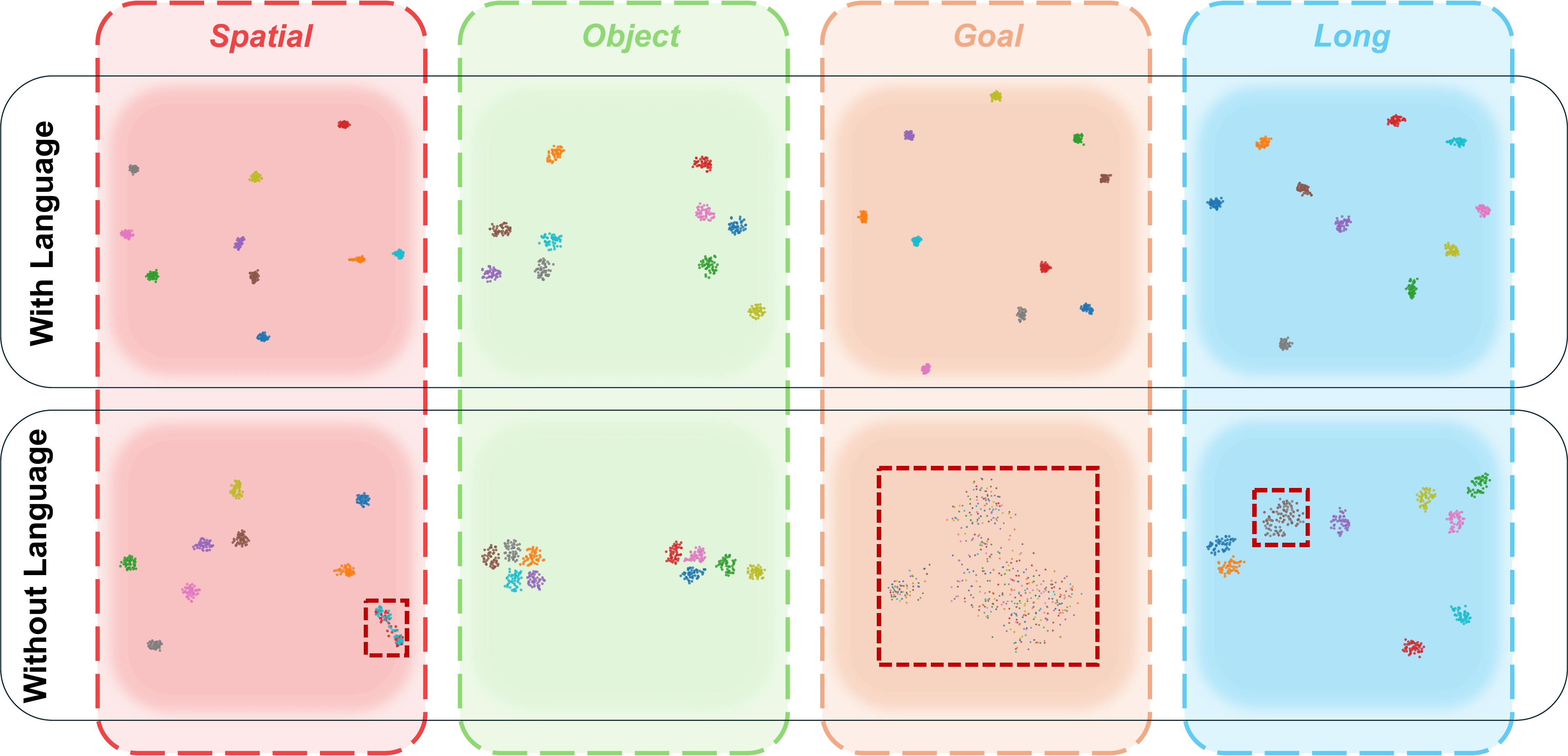}
    \caption{
    \textbf{t-SNE of visual-prefix representations on LIBERO, with and without language.}
    Colors denote tasks. With language (top), each suite forms ten well-separated task clusters.
    Without language (bottom), tasks stay separable only where the scene identifies them:
    Object remains separable, Spatial and Long each merge a few tasks (red dashed boxes),
    and Goal, whose tasks share one scene, collapses into a single cloud,
    mirroring the task-head accuracy in Table~\ref{tab:lang_cond_libero_comparison}.
}
    \label{fig:lang-conditioning-tsne-full}
\end{figure}

\begin{figure}[t]
    \centering
    \includegraphics[width=1.0\textwidth]{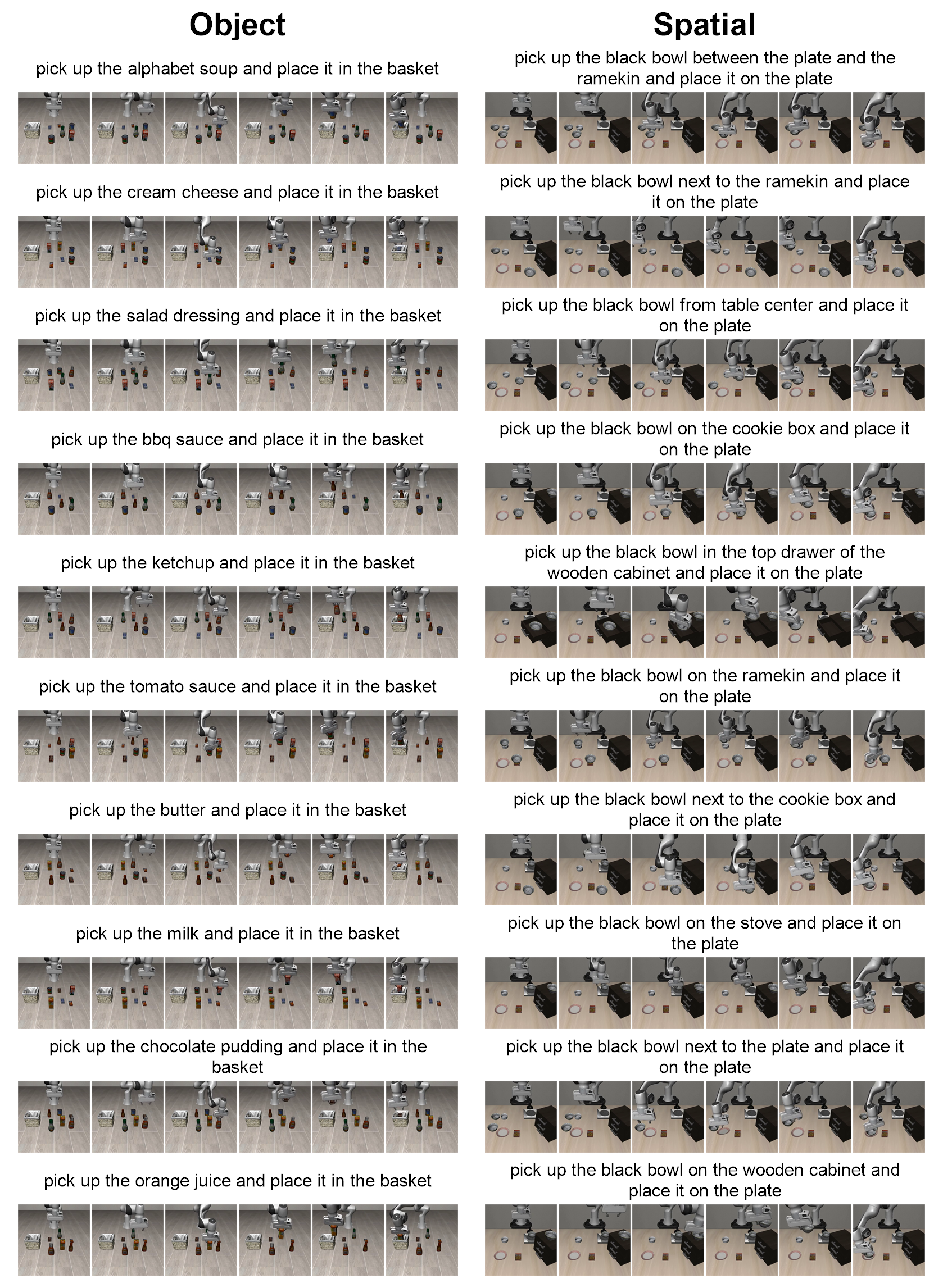}
    \caption{
    \textbf{Example CoRP rollouts on LIBERO-Object and LIBERO-Spatial.}
    Each row shows six frames from one rollout of the task named above it, from left to right.
    Object tasks require picking the named item and placing it in the basket;
    Spatial tasks move the same black bowl to the plate from different starting locations.
}
    \label{fig:libero1}
\end{figure}

\begin{figure}[t]
    \centering
    \includegraphics[width=1.0\textwidth]{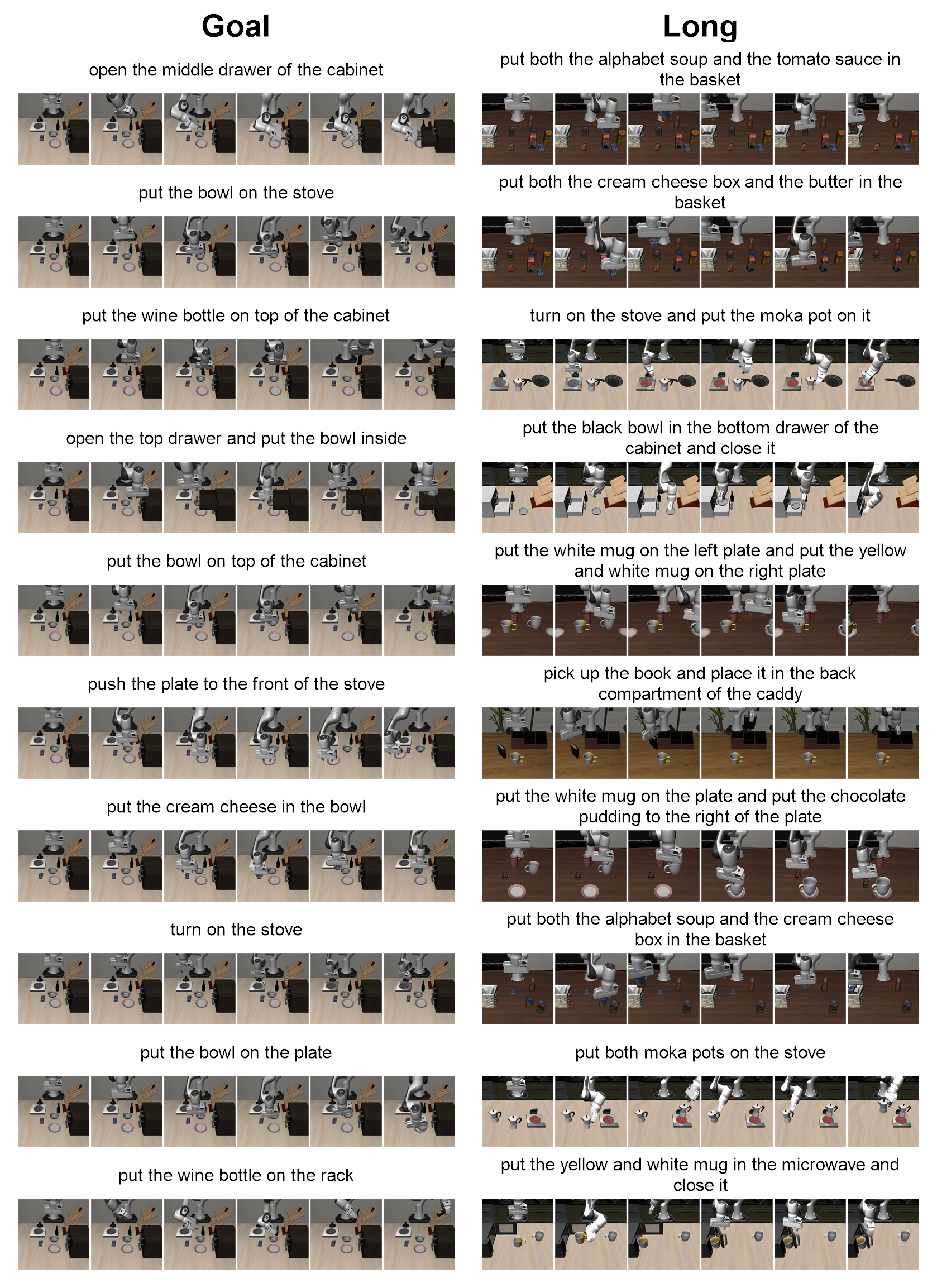}
    \caption{
    \textbf{Example CoRP rollouts on LIBERO-Goal and LIBERO-Long.}
    Each row shows six frames from one rollout of the task named above it, from left to right.
    Goal tasks share a single scene and differ only in the instruction;
    Long tasks are multi-step, most chaining two subgoals.
}
    \label{fig:libero2}
\end{figure}

\paragraph{LIBERO.}
We train a single joint model across four suites with 40 tasks.
We do not split the dataset for validation, so we uniformly sample all data for training.
We use the chunk length $H=12$ for LIBERO.
If fewer than $H$ actions remain, we repeat the final action and mark the repeated positions as invalid.
Invalid positions do not contribute to the flow-matching loss $\mathcal L_{fm}$
Figure~\ref{fig:lang-conditioning-tsne-full} shows t-SNE analysis for all 40 tasks, with and without language conditioning.
We follow the original LIBERO evaluation protocol to identify whether the task is successful.
We repeated every task 50 times with randomized initial conditions, for a total of 2,000 trials.
Figures~\ref{fig:libero1} and~\ref{fig:libero2} record the manipulation process. 

\paragraph{RoboTwin~2.0.}
In all experiments, we report a model trained on 50 tasks.
We use the chunk length $H=50$ for RoboTwin~2.0.
We also have no validation data, and we feed all data into the training process.

\begin{figure}[p]
    \centering
    \includegraphics[height=0.95\textheight]{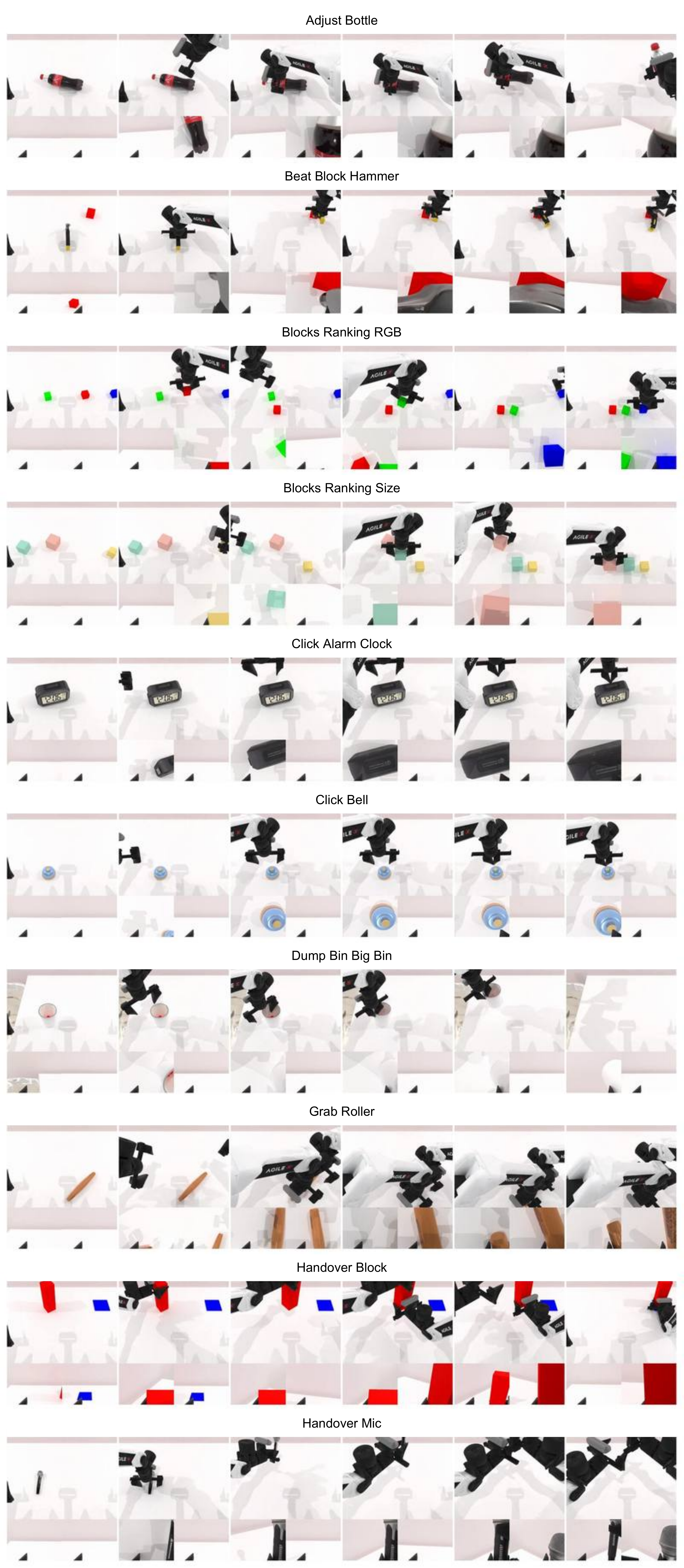}
    \caption{
    \textbf{Example CoRP rollouts on RoboTwin~2.0 (\textit{Clean}), tasks 1--10.}
    Each row shows six frames from one rollout, from left to right in time;
    each frame combines the head-camera view (top) with the two wrist-camera views (bottom).
    Figures~\ref{fig:robotwin1}--\ref{fig:robotwin5} cover all 50 tasks in alphabetical order.}
    \label{fig:robotwin1}
\end{figure}

\begin{figure}[p]
    \centering
    \includegraphics[height=0.95\textheight]{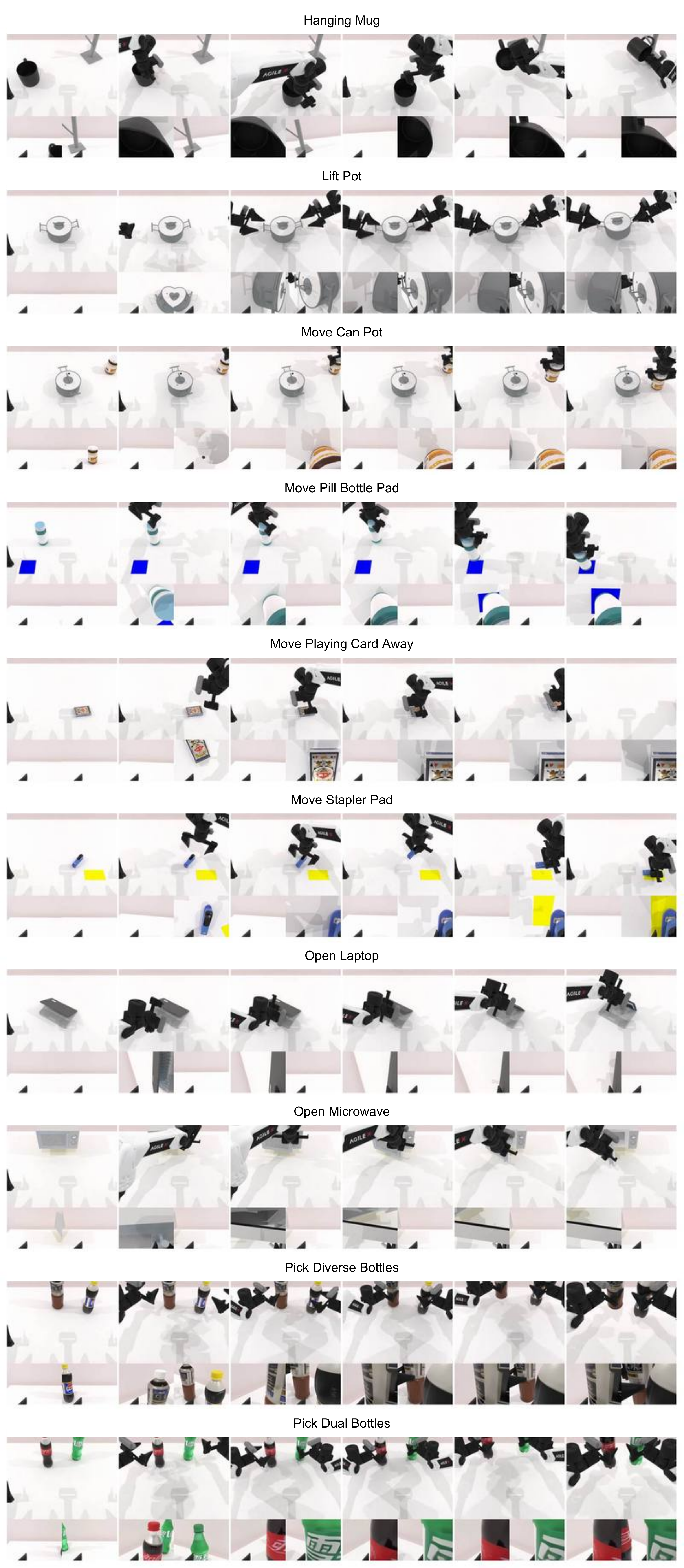}
    \caption{
    \textbf{Example CoRP rollouts on RoboTwin~2.0 (\textit{Clean}), tasks 11--20.}
    Layout as in Figure~\ref{fig:robotwin1}.}
    \label{fig:robotwin2}
\end{figure}

\begin{figure}[p]
    \centering
    \includegraphics[height=0.95\textheight]{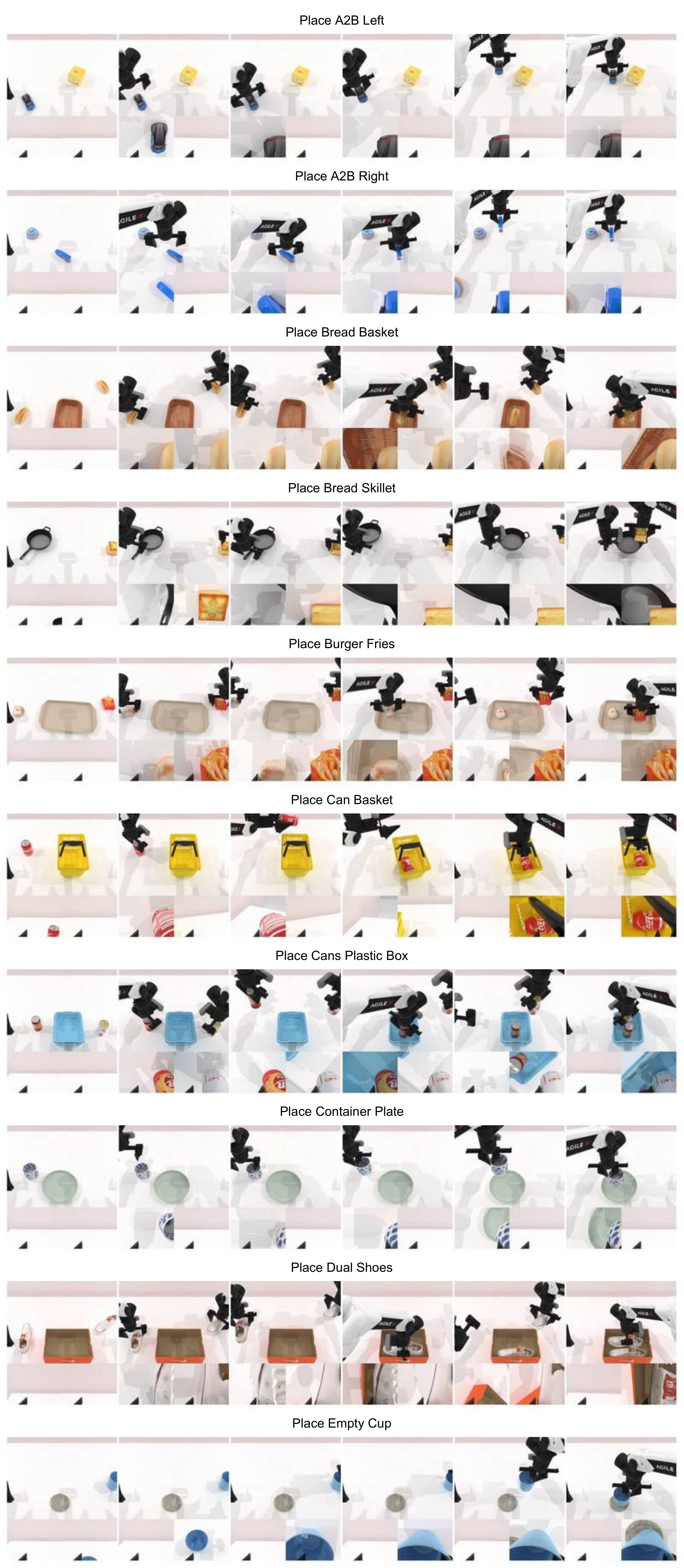}
    \caption{
    \textbf{Example CoRP rollouts on RoboTwin~2.0 (\textit{Clean}), tasks 21--30.}
    Layout as in Figure~\ref{fig:robotwin1}.}
    \label{fig:robotwin3}
\end{figure}

\begin{figure}[p]
    \centering
    \includegraphics[height=0.95\textheight]{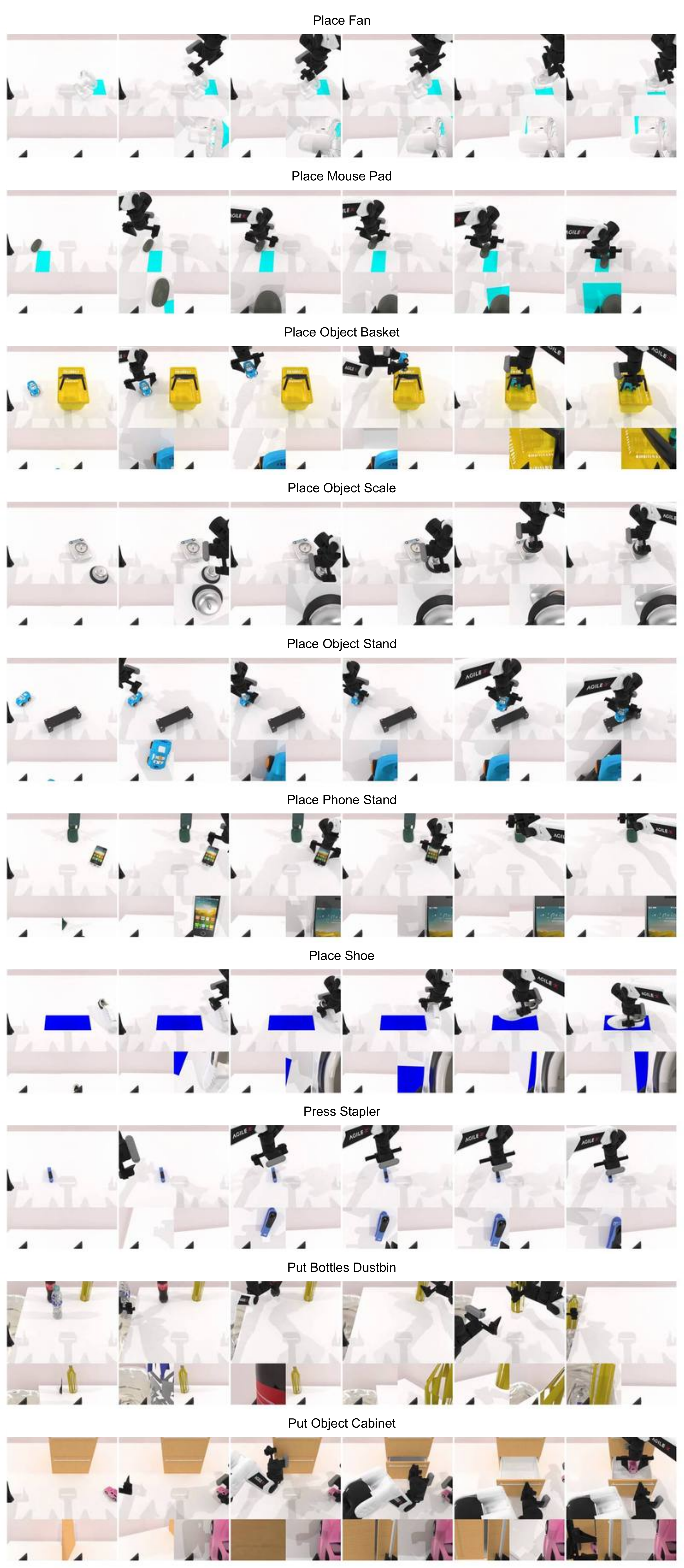}
    \caption{
    \textbf{Example CoRP rollouts on RoboTwin~2.0 (\textit{Clean}), tasks 31--40.}
    Layout as in Figure~\ref{fig:robotwin1}.}
    \label{fig:robotwin4}
\end{figure}

\begin{figure}[p]
    \centering
    \includegraphics[height=0.95\textheight]{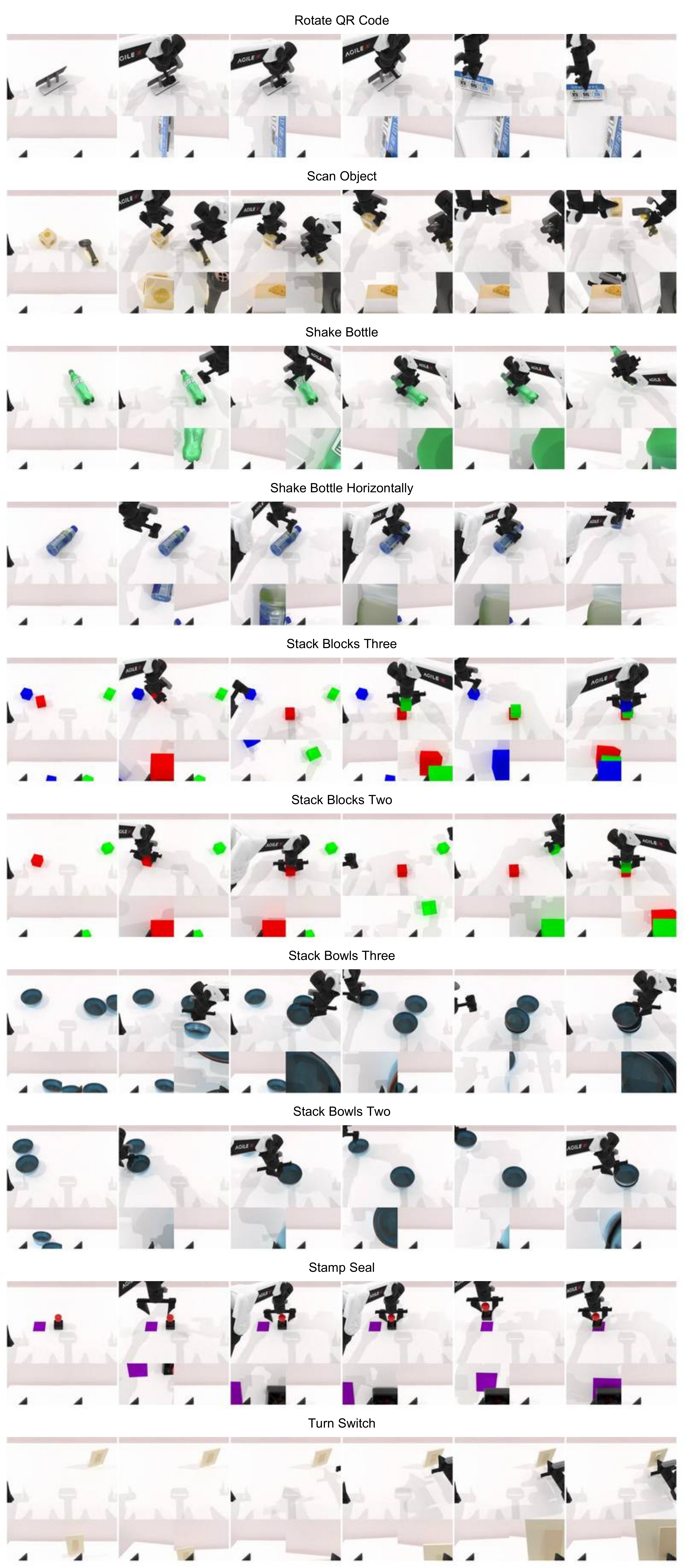}
    \caption{
    \textbf{Example CoRP rollouts on RoboTwin~2.0 (\textit{Clean}), tasks 41--50.}
    Layout as in Figure~\ref{fig:robotwin1}.}
    \label{fig:robotwin5}
\end{figure}

\begin{figure}[p]
    \centering
    \includegraphics[height=0.95\textheight]{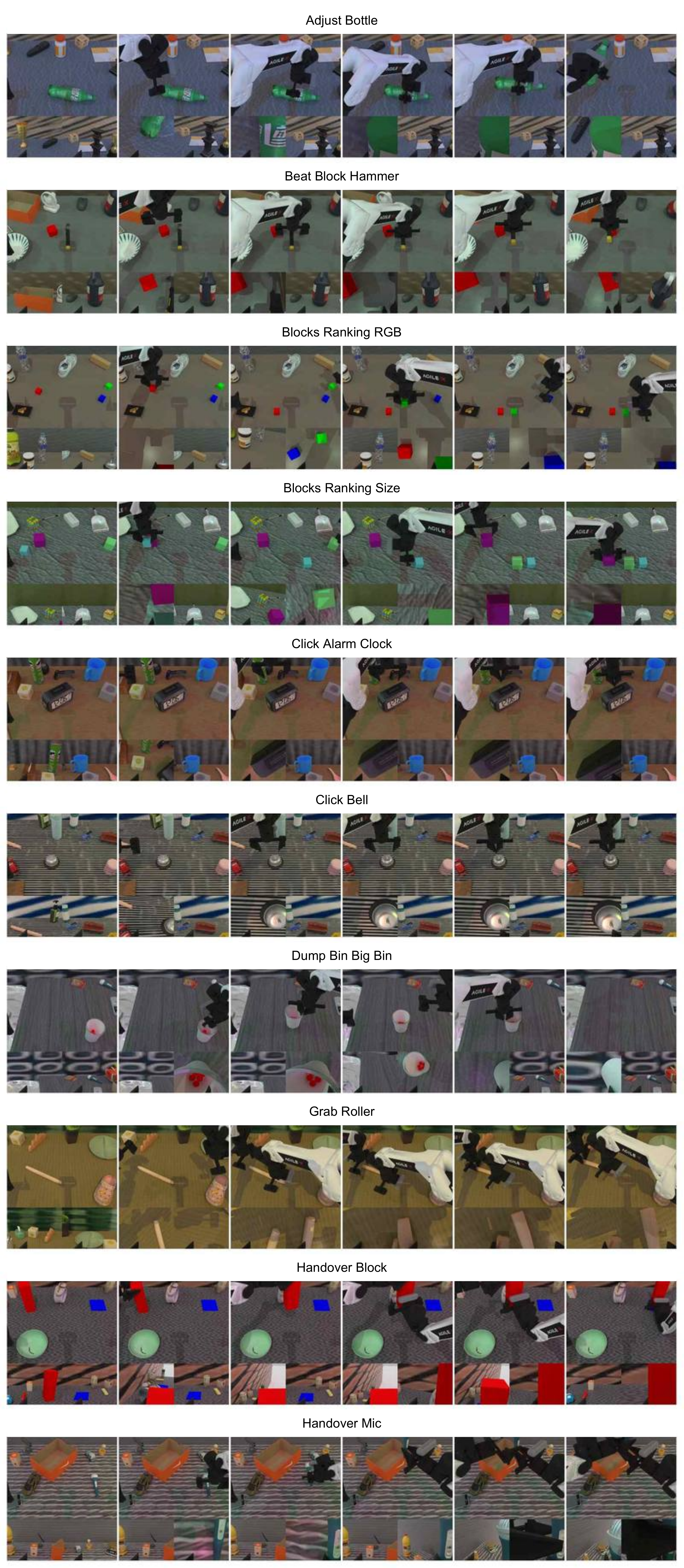}
    \caption{
    \textbf{Example CoRP rollouts on RoboTwin~2.0 (\textit{Randomized}), tasks 1--10.}
    Same tasks and layout as Figure~\ref{fig:robotwin1}, under randomized distractors, textures, and lighting.
    Figures~\ref{fig:robotwin6}--\ref{fig:robotwin10} cover all 50 tasks in alphabetical order.}
    \label{fig:robotwin6}
\end{figure}

\begin{figure}[p]
    \centering
    \includegraphics[height=0.95\textheight]{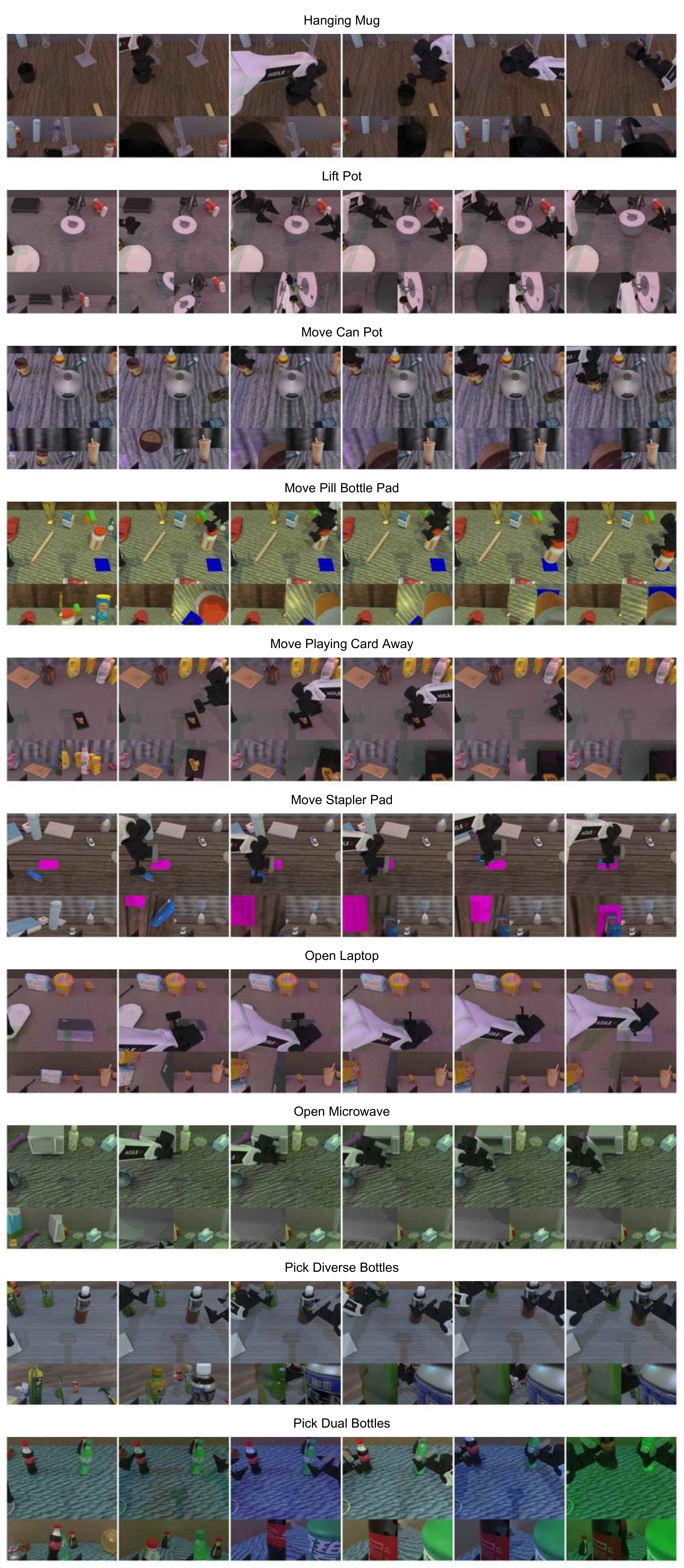}
    \caption{
    \textbf{Example CoRP rollouts on RoboTwin~2.0 (\textit{Randomized}), tasks 11--20.}
    Same tasks and layout as Figure~\ref{fig:robotwin2}.}
    \label{fig:robotwin7}
\end{figure}

\begin{figure}[p]
    \centering
    \includegraphics[height=0.95\textheight]{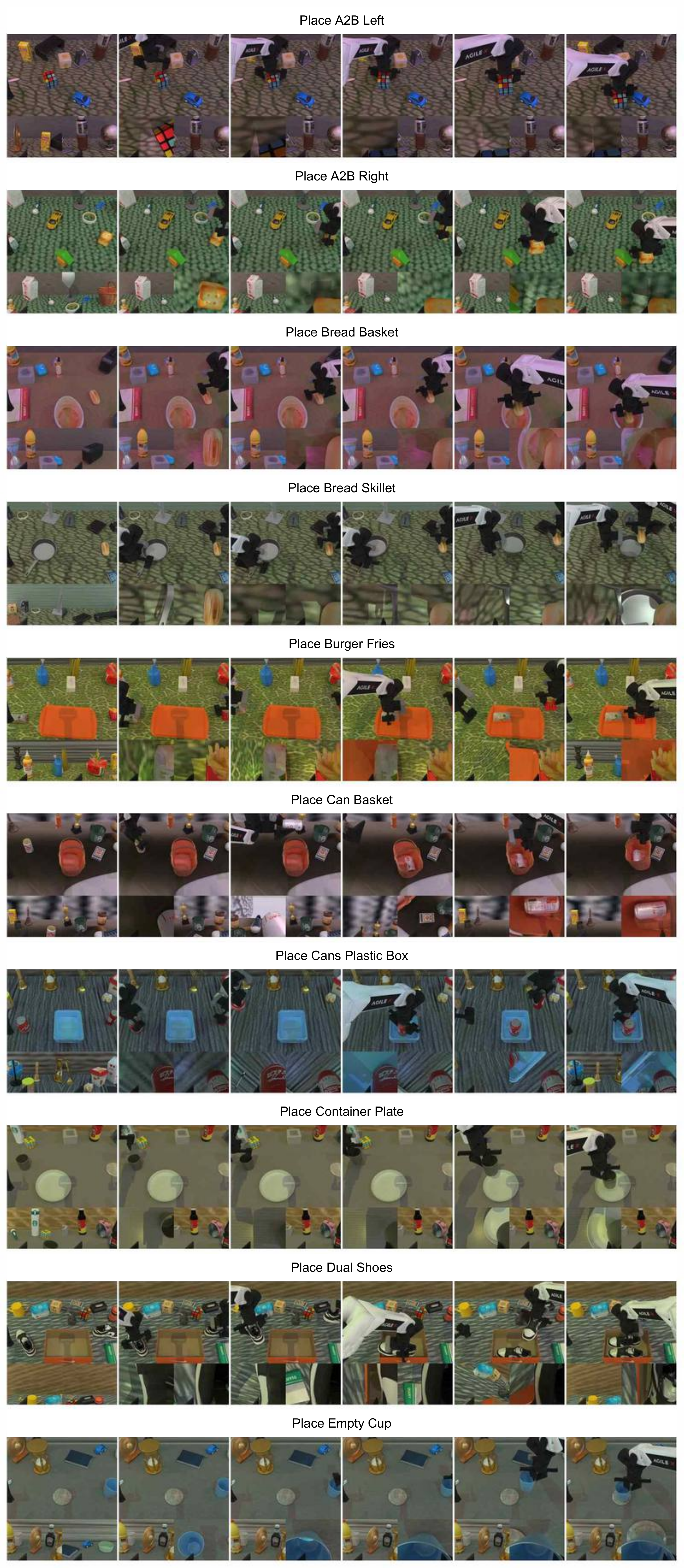}
    \caption{
    \textbf{Example CoRP rollouts on RoboTwin~2.0 (\textit{Randomized}), tasks 21--30.}
    Same tasks and layout as Figure~\ref{fig:robotwin3}.}
    \label{fig:robotwin8}
\end{figure}

\begin{figure}[p]
    \centering
    \includegraphics[height=0.95\textheight]{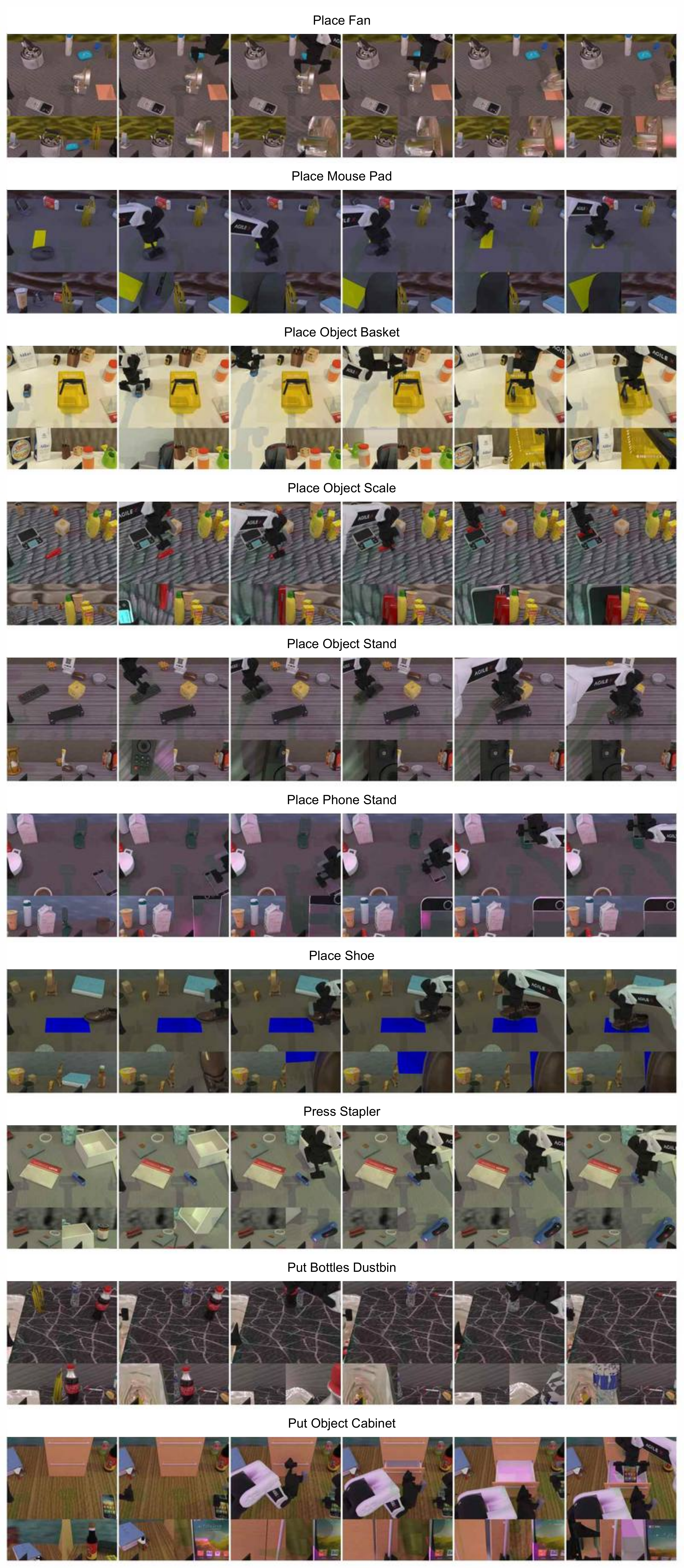}
    \caption{
    \textbf{Example CoRP rollouts on RoboTwin~2.0 (\textit{Randomized}), tasks 31--40.}
    Same tasks and layout as Figure~\ref{fig:robotwin4}.}
    \label{fig:robotwin9}
\end{figure}

\begin{figure}[p]
    \centering
    \includegraphics[height=0.95\textheight]{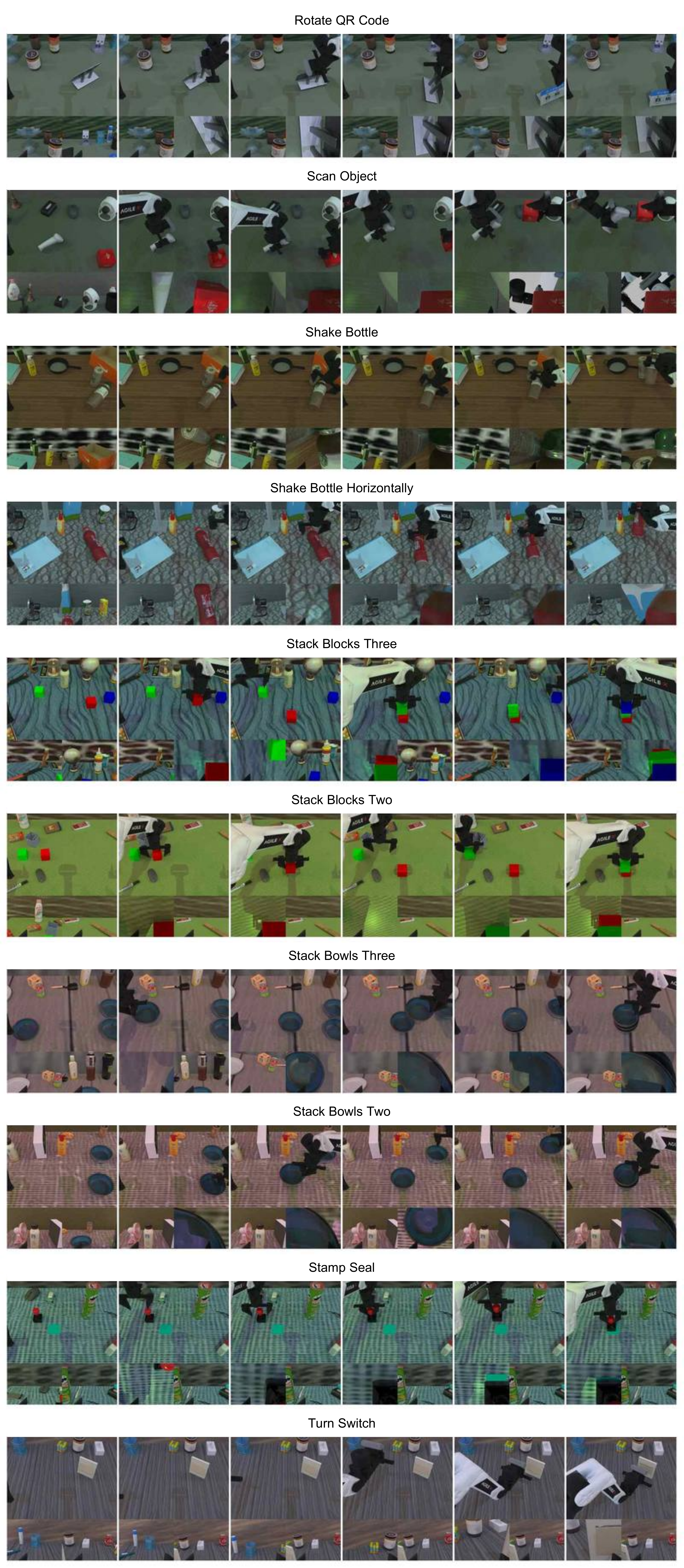}
    \caption{
    \textbf{Example CoRP rollouts on RoboTwin~2.0 (\textit{Randomized}), tasks 41--50.}
    Same tasks and layout as Figure~\ref{fig:robotwin5}.}
    \label{fig:robotwin10}
\end{figure}

We report detailed per-task evaluation on the RoboTwin~2.0 benchmark in \textit{Clean} and \textit{Randomized} settings.
These two settings differ only in evaluation conditions, while the training setup remains identical.
We follow the original evaluation protocol to assess task success.
We repeated every task 100 times for evaluation, for a total of $10,000$ trials.
The rollouts for different tasks are presented in Figures~\ref{fig:robotwin1}--~\ref{fig:robotwin10}

\begin{table}[t]
  \caption{\textbf{Task-wise success rates (\%) on RoboTwin 2.0.}
  Best results are in bold for each task and setting, including ties.}
  \label{tab:robotwin-full}
  \centering
  \begingroup
  \setlength{\tabcolsep}{3pt}
  \renewcommand{\arraystretch}{0.96}
  \resizebox{\linewidth}{!}{%
    \begin{tabular}{@{}l*{8}{cc}@{}}
      \toprule
      & \multicolumn{2}{c}{$\pi_0$}
      & \multicolumn{2}{c}{$\pi_{0.5}$}
      & \multicolumn{2}{c}{X-VLA}
      & \multicolumn{2}{c}{Motus}
      & \multicolumn{2}{c}{GWP}
      & \multicolumn{2}{c}{DP}
      & \multicolumn{2}{c}{ACT}
      & \multicolumn{2}{c}{Ours} \\
      \cmidrule(lr){2-3}
      \cmidrule(lr){4-5}
      \cmidrule(lr){6-7}
      \cmidrule(lr){8-9}
      \cmidrule(lr){10-11}
      \cmidrule(lr){12-13}
      \cmidrule(lr){14-15}
      \cmidrule(lr){16-17}
      Task
      & C & R
      & C & R
      & C & R
      & C & R
      & C & R
      & C & R
      & C & R
      & C & R \\
      \midrule

      Adjust Bottle
      & 90 & 56 & 79 & 83 & \textbf{100} & 99
      & 89 & 93 & \textbf{100} & \textbf{100}
      & 97 & 0 & 97 & 23 & 98 & 98 \\

      Beat Block Hammer
      & 43 & 21 & 63 & 50 & 92 & 88
      & 95 & 88 & 86 & 86
      & 42 & 0 & 56 & 3 & \textbf{97} & \textbf{89} \\

      Blocks Ranking RGB
      & 19 & 5 & 43 & 35 & 83 & 83
      & \textbf{99} & \textbf{97} & 92 & 96
      & 0 & 0 & 1 & 0 & 55 & 61 \\

      Blocks Ranking Size
      & 7 & 1 & 8 & 14 & 67 & \textbf{74}
      & \textbf{75} & 63 & 44 & 48
      & 1 & 0 & 0 & 0 & 48 & 42 \\

      Click Alarm clock
      & 63 & 11 & 97 & 93 & 99 & 99
      & \textbf{100} & \textbf{100} & \textbf{100} & \textbf{100}
      & 61 & 5 & 32 & 4 & 97 & 92 \\

      Click Bell
      & 44 & 3 & 75 & 76 & \textbf{100} & \textbf{100}
      & \textbf{100} & \textbf{100} & \textbf{100} & \textbf{100}
      & 54 & 0 & 58 & 3 & \textbf{100} & 99 \\

      Dump Bin Bigbin
      & 83 & 24 & 30 & 42 & 79 & 77
      & \textbf{95} & 91 & 92 & \textbf{100}
      & 49 & 0 & 68 & 1 & 70 & 70 \\

      Grab Roller
      & 96 & 80 & 90 & 89 & \textbf{100} & \textbf{100}
      & \textbf{100} & \textbf{100} & \textbf{100} & \textbf{100}
      & 98 & 0 & 94 & 25 & \textbf{100} & 99 \\

      Handover Block
      & 45 & 8 & 18 & 19 & 73 & 37
      & \textbf{86} & 73 & 80 & \textbf{80}
      & 10 & 0 & 42 & 0 & 79 & 50 \\

      Handover Mic
      & 98 & 13 & 28 & 18 & 0 & 0
      & 78 & 63 & 72 & 72
      & 53 & 0 & 85 & 0 & \textbf{99} & \textbf{96} \\

      Hanging Mug
      & 11 & 3 & 3 & 3 & 23 & 27
      & \textbf{38} & \textbf{38} & 16 & 12
      & 8 & 0 & 7 & 0 & 1 & 8 \\

      Lift Pot
      & 84 & 36 & 0 & 0 & 99 & \textbf{100}
      & 96 & 99 & 98 & 98
      & 39 & 0 & 88 & 0 & \textbf{100} & 98 \\

      Move Can Pot
      & 58 & 21 & 29 & 27 & \textbf{89} & \textbf{86}
      & 34 & 74 & 76 & 78
      & 39 & 0 & 22 & 4 & 85 & 75 \\

      Move Pillbottle Pad
      & 21 & 1 & 33 & 29 & 73 & 71
      & \textbf{93} & \textbf{96} & 90 & 90
      & 1 & 0 & 0 & 0 & 88 & 74 \\

      Move Playingcard Away
      & 53 & 22 & 59 & 67 & 93 & \textbf{98}
      & \textbf{100} & 96 & 78 & 72
      & 47 & 0 & 36 & 0 & 96 & 96 \\

      Move Stapler Pad
      & 0 & 2 & 16 & 18 & 78 & 73
      & 83 & \textbf{85} & \textbf{92} & 82
      & 1 & 0 & 0 & 0 & 42 & 37 \\

      Open Laptop
      & 85 & 46 & 19 & 35 & 93 & \textbf{100}
      & 95 & 91 & 96 & 98
      & 49 & 0 & 56 & 0 & \textbf{99} & 96 \\

      Open Microwave
      & 80 & 50 & 35 & 37 & 79 & 71
      & \textbf{95} & \textbf{91} & 74 & 66
      & 5 & 0 & 86 & 0 & 4 & 15 \\

      Pick Diverse Bottles
      & 27 & 6 & 5 & 3 & 58 & 36
      & 90 & \textbf{91} & 82 & 70
      & 6 & 0 & 7 & 0 & \textbf{93} & 75 \\

      Pick Dual Bottles
      & 57 & 12 & 10 & 6 & 47 & 36
      & 96 & \textbf{90} & 86 & 86
      & 24 & 0 & 31 & 0 & \textbf{97} & 80 \\

      Place A2b Left
      & 31 & 1 & 62 & 60 & 48 & 49
      & 88 & 79 & \textbf{94} & \textbf{88}
      & 2 & 0 & 1 & 0 & 59 & 60 \\

      Place A2b Right
      & 27 & 6 & 62 & 57 & 36 & 36
      & \textbf{91} & 87 & 90 & \textbf{92}
      & 13 & 0 & 0 & 0 & 63 & 67 \\

      Place Bread Basket
      & 17 & 4 & 48 & 56 & 81 & 71
      & \textbf{91} & \textbf{94} & 82 & 82
      & 14 & 0 & 6 & 0 & 78 & 78 \\

      Place Bread Skillet
      & 23 & 1 & 38 & 46 & 77 & 67
      & 86 & 83 & \textbf{94} & 90
      & 11 & 0 & 7 & 0 & 92 & \textbf{95} \\

      Place Burger Fries
      & 80 & 4 & 66 & 70 & 94 & 94
      & \textbf{98} & \textbf{98} & \textbf{98} & 96
      & 72 & 0 & 49 & 0 & 96 & 97 \\

      Place Can Basket
      & 41 & 5 & 19 & 25 & 49 & 52
      & \textbf{81} & \textbf{76} & 78 & 74
      & 18 & 0 & 1 & 0 & 70 & 64 \\

      Place Cans Plastic box
      & 34 & 2 & 40 & 47 & 97 & 98
      & 98 & 94 & \textbf{100} & \textbf{100}
      & 40 & 0 & 16 & 0 & 97 & 95 \\

      Place Container Plate
      & 88 & 45 & 71 & 78 & 97 & 95
      & \textbf{98} & \textbf{96} & \textbf{98} & \textbf{96}
      & 41 & 0 & 72 & 1 & 85 & 85 \\

      Place Dual Shoes
      & 15 & 0 & 12 & 7 & 79 & \textbf{88}
      & 93 & 87 & \textbf{96} & 84
      & 8 & 0 & 9 & 0 & 77 & 81 \\

      Place Empty Cup
      & 37 & 11 & 75 & 86 & \textbf{100} & \textbf{98}
      & 99 & \textbf{98} & 90 & 90
      & 37 & 0 & 61 & 0 & 91 & 86 \\

      Place Fan
      & 20 & 10 & 25 & 36 & 80 & 75
      & 91 & 87 & \textbf{92} & \textbf{94}
      & 3 & 0 & 1 & 0 & 79 & 78 \\

      Place Mouse Pad
      & 7 & 1 & 21 & 26 & 70 & 70
      & 66 & 68 & \textbf{88} & \textbf{90}
      & 0 & 0 & 0 & 0 & 57 & 58 \\

      Place Object Basket
      & 16 & 2 & 43 & 36 & 44 & 39
      & 81 & 87 & \textbf{90} & \textbf{92}
      & 15 & 0 & 15 & 0 & 79 & 69 \\

      Place Object Scale
      & 10 & 0 & 40 & 49 & 52 & 74
      & \textbf{88} & \textbf{85} & \textbf{88} & 80
      & 1 & 0 & 0 & 0 & 82 & 70 \\

      Place Object Stand
      & 36 & 11 & 74 & 65 & 86 & 88
      & 98 & 97 & \textbf{100} & \textbf{98}
      & 22 & 0 & 1 & 0 & 85 & 83 \\

      Place Phone Stand
      & 35 & 7 & 49 & 53 & \textbf{88} & 87
      & 87 & 86 & 82 & 72
      & 13 & 0 & 2 & 0 & 80 & \textbf{88} \\

      Place Shoe
      & 28 & 6 & 57 & 61 & 96 & 95
      & \textbf{99} & \textbf{97} & 98 & 96
      & 23 & 0 & 5 & 0 & 93 & 95 \\

      Press Stapler
      & 62 & 29 & 80 & 70 & 92 & \textbf{98}
      & 93 & \textbf{98} & \textbf{96} & 96
      & 6 & 0 & 31 & 6 & \textbf{96} & 97 \\

      Put Bottles in Dustbin
      & 54 & 13 & 12 & 9 & 74 & 77
      & \textbf{81} & \textbf{79} & 72 & 70
      & 22 & 0 & 27 & 1 & 20 & 14 \\

      Put Object Cabinet
      & 68 & 18 & 24 & 15 & 46 & 48
      & \textbf{88} & 71 & 74 & \textbf{74}
      & 42 & 0 & 15 & 0 & 53 & 57 \\

      Rotate QR code
      & 68 & 15 & 47 & 56 & 34 & 33
      & 89 & 73 & \textbf{90} & 84
      & 13 & 0 & 1 & 0 & 82 & \textbf{85} \\

      Scan Object
      & 18 & 1 & 42 & 38 & 14 & 36
      & 67 & 66 & 60 & 64
      & 9 & 0 & 2 & 0 & \textbf{85} & \textbf{80} \\

      Shake Bottle
      & 97 & 60 & 91 & \textbf{100} & 99 & \textbf{100}
      & \textbf{100} & 97 & \textbf{100} & \textbf{100}
      & 65 & 8 & 74 & 10 & 99 & 99 \\

      Shake Bottle Horizontally
      & 99 & 51 & 96 & \textbf{100} & \textbf{100} & \textbf{100}
      & \textbf{100} & 98 & \textbf{100} & 98
      & 59 & 18 & 63 & 4 & 97 & 97 \\

      Stack Blocks Three
      & 17 & 0 & 15 & 16 & 6 & 10
      & \textbf{91} & \textbf{95} & 70 & 78
      & 0 & 0 & 0 & 0 & 16 & 30 \\

      Stack Blocks Two
      & 42 & 1 & 48 & 56 & 92 & 87
      & 99 & \textbf{98} & \textbf{100} & 94
      & 7 & 0 & 25 & 0 & 71 & 71 \\

      Stack Bowls Three
      & 66 & 24 & 33 & 35 & 76 & 86
      & \textbf{79} & \textbf{87} & 70 & 72
      & 63 & 0 & 48 & 0 & 53 & 52 \\

      Stack Bowls Two
      & 91 & 41 & 78 & 66 & 96 & 93
      & \textbf{98} & \textbf{98} & 96 & 92
      & 61 & 0 & 82 & 0 & 71 & 78 \\

      Stamp Seal
      & 3 & 4 & 36 & 23 & 76 & 82
      & 93 & 92 & \textbf{96} & \textbf{98}
      & 2 & 0 & 2 & 0 & 82 & 55 \\

      Turn Switch
      & 27 & 23 & 5 & 6 & 40 & 61
      & \textbf{84} & 78 & 82 & \textbf{84}
      & 36 & 1 & 5 & 2 & 53 & 54 \\

      \midrule
      Average
      & 46.42 & 16.34
      & 42.98 & 43.84
      & 72.88 & 72.84
      & \textbf{88.64} & \textbf{86.96}
      & 86.36 & 85.04
      & 28.04 & 0.64
      & 29.74 & 1.74
      & 75.78 & 73.36 \\
      \bottomrule
    \end{tabular}%
  }
  \par
  \endgroup
\end{table}

\begin{figure}[h]
    \centering
    \includegraphics[width=0.8\textwidth]{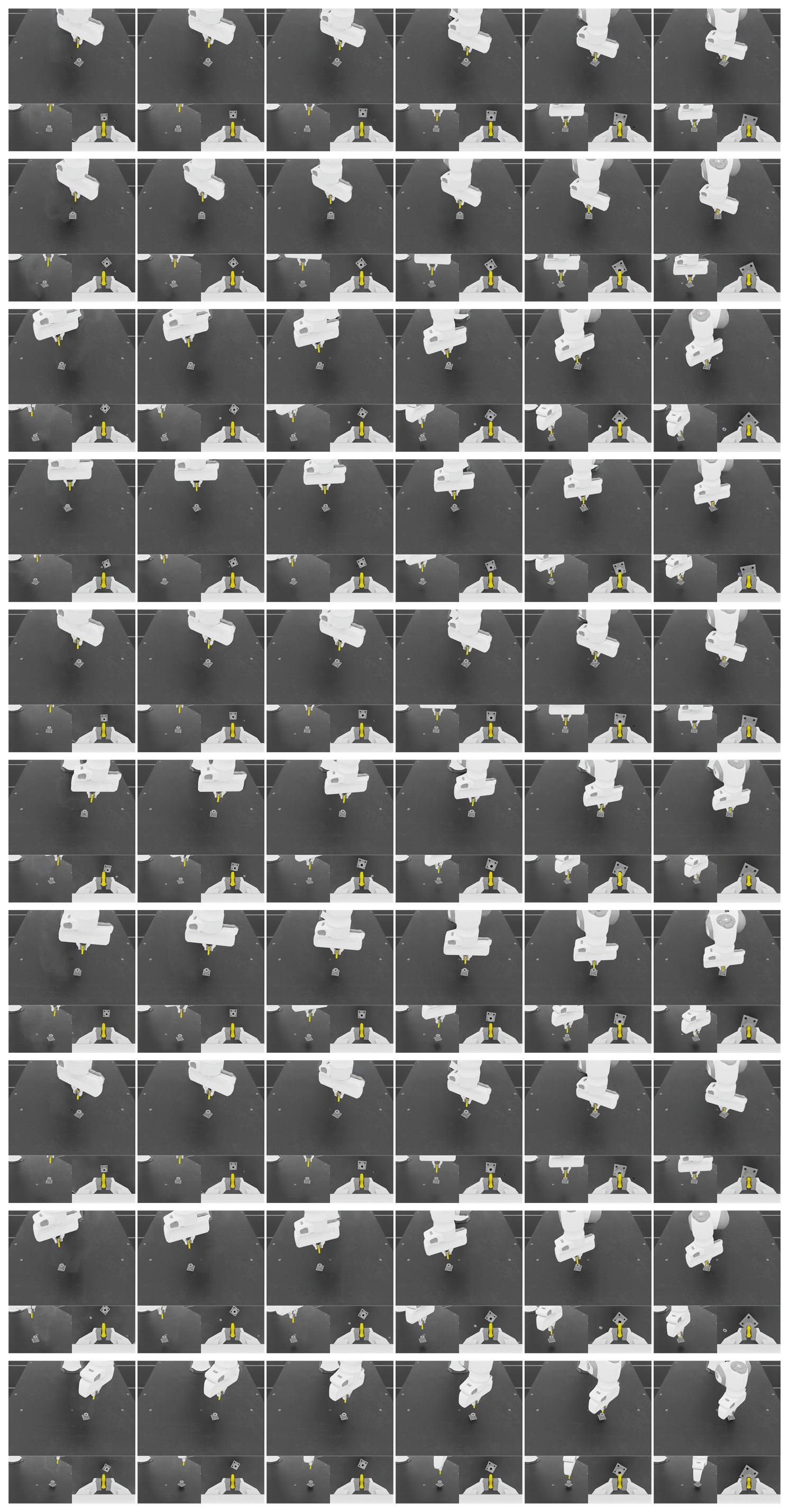}
    \caption{
    \textbf{Example CoRP rollouts on the Forge peg-insertion task.}
    Each row shows six frames from one rollout with a different initial condition, from left to right in time;
    Each frame combines the main view (top) with two auxiliary views (bottom).
    The policy uses vision only to insert an 8.0\,mm peg into an 8.1\,mm hole, reaching 84\% success over 50 trials.
}
    \label{fig:peg_insert}
\end{figure}

\paragraph{Forge~\citep{noseworthy2025forge}.}
We have an additional simulation for precision manipulation. We choose the task \textit{peg insertion} and adopt only visual information. The task is difficult because it requires precise peg manipulation. The hole has an internal diameter of $8.1 \,\mathrm{ms}$, while the peg has a diameter of $8.0\,\mathrm{ms}$.
We train only 30k steps from scratch for this task. However, we achieved an $84\%$ success rate over 50 trials, similar to the best result reported in \citep{noseworthy2025forge}. As visualized in Figure~\ref{fig:peg_insert}, the robot smoothly inserts the peg into the hole from different initial conditions. In this task, we use $H=12$, and the control frequency is $ 15 \, \mathrm {Hz} $. We keep the other configurations the same as in the aforementioned simulations.

\subsection{Baseline Policies}\label{app:baselines}
We summarize the baseline policies compared in Tables~\ref{tab:libero_comparison} and~\ref{tab:robotwin_comparison}. 
We organize them into three groups according to their use of pretrained vision–language or video-generation backbones. 
We include methods that combine both types of components in the video-model-based group. 
This grouping concerns backbone initialization rather than whether a policy accepts language inputs or uses pretraining in general.

\subsubsection{Policies based on pretrained vision–language models}\label{app:baselines1}
\textbf{OpenVLA}~\citep{kim2024openvla} adapts the vision–language backbone for robot action prediction. 
It combines a Llama 2 language model with visual features from DINOv2 and SigLIP, and is trained on heterogeneous robot demonstrations. 
Each continuous action dimension is quantized into discrete tokens, allowing the model to predict robot commands autoregressively from image observations and language instructions using a next-token prediction objective. 

\textbf{SmolVLA}~\citep{shukor2025smolvla} combines a SmolVLM-2 backbone with a flow-matching action expert. 
The expert alternates self-attention over action tokens with cross-attention to multimodal features to generate continuous action chunks. 
Visual-token reduction and intermediate VLM layers reduce computational cost, while an asynchronous inference scheme can decouple action generation from execution. 
Our LIBERO comparison includes the 2.25B configuration reported in the original paper. 

{$\mathbf{\pi_0}$}~\citep{black2025pi0} couples a pretrained PaliGemma vision–language model with a dedicated action expert trained using flow matching. 
Images, language instructions, and proprioceptive states condition the generation of continuous action chunks, avoiding the need to discretize control outputs. 
Cross-embodiment robot pretraining combines general visual–language representations with motor supervision, after which the policy can be adapted to downstream manipulation tasks through task-specific post-training. 

{$\mathbf{\pi_{0.5}}$}~\citep{black2025pi05} extends the $\pi_0$ framework with a broader training mixture that includes robot demonstrations, semantic prediction tasks, and non-robot vision–language data. 
Its hybrid training recipe uses discrete action tokens during pretraining and introduces a flow-matching action expert during post-training for continuous control. 
In its original formulation, the model can also predict intermediate language subtasks and condition action generation on them, connecting high-level task decomposition with low-level execution. 

\textbf{OpenVLA-OFT}~\citep{kim2025openvlaoft} applies an optimized fine-tuning recipe to OpenVLA. 
Instead of autoregressively decoding discretized actions, it predicts continuous action chunks in parallel and trains the action outputs with an $L_1$ regression objective. 
The resulting policy also accommodates additional inputs such as multiple camera views and proprioceptive states. 
It therefore provides a comparison to VLM-based control in which action decoding and fine-tuning are optimized without introducing an iterative diffusion or flow-matching action head. 

\textbf{X-VLA}~\citep{zheng2026xvla} uses learnable domain-specific soft prompts to accommodate differences in robot embodiments and data-collection setups. 
A Florence-2 encoder processes the main image–language stream, while auxiliary camera views are encoded separately. 
These representations are combined with proprioceptive inputs and soft prompts in a shared Transformer for flow-matching action generation. 
The prompts provide domain-dependent conditioning without requiring a separate policy backbone for every embodiment. 

\subsubsection{Policies based on pretrained video-generation models}\label{app:baselines2}
\textbf{Motus}~\citep{bi2026motus} integrates understanding, video-generation, and action experts through a Mixture-of-Transformers architecture. 
It learns latent actions from optical flow to capture motion information that can transfer across embodiments. 
Modality-specific diffusion timesteps allow the model to switch among action prediction, video prediction, inverse dynamics, and joint video–action modeling. 
Its staged training transfers general multimodal priors and motion knowledge from heterogeneous datasets to robot control. 

\textbf{Fast-WAM-Joint}~\citep{yuan2026fastwam} is the joint-generation variant investigated within the Fast-WAM framework. 
It couples a pretrained Wan2.2 video diffusion Transformer with an action expert and allows attention between future-video and action tokens during joint denoising.
Unlike the main Fast-WAM model, this variant retains future-video generation at inference time. 
It therefore represents coupled video–action prediction rather than the action-only inference procedure used by Fast-WAM.

\textbf{Cosmos-Policy}~\citep{kim2026cosmospolicy} adapts a pretrained video diffusion model to robot control through latent-frame injection. 
Robot states, action chunks, and value targets are represented in the existing latent-frame format, allowing the denoising backbone to learn action prediction, future-state prediction, and value estimation without a separate action-specific network.
The framework supports direct policy execution and, after refinement using rollout data, value-guided action selection. 
The LIBERO results in our comparison correspond to direct policy evaluation rather than test-time planning.

\textbf{GigaWorld-Policy}~\citep{giga2026gigaworld} initializes a unified policy from a pretrained video-generation backbone and trains it to predict action chunks together with action-conditioned future observations. 
Its blockwise causal attention mask prevents action tokens from attending to future-video tokens, while allowing future-video prediction to depend on the actions. 
This asymmetric information flow makes future prediction an auxiliary training objective rather than a prerequisite for action generation, allowing future-video synthesis to be omitted during inference.

\subsubsection{Policies without either pretrained backbone}\label{app:baselines3}
\textbf{ACT}~\citep{zhao2023act} is a Transformer-based imitation learning method that models action chunks using a conditional variational autoencoder. 
Given camera observations and robot joint states, it predicts a sequence of future actions instead of a single command. 
A latent variable captures variation in demonstrations during training, while temporal ensembling can combine overlapping predictions during execution to produce smoother control. 

\textbf{Diffusion Policy}~\citep{chi2023diffusionpolicy} models a visuomotor policy as a conditional denoising process over action sequences. 
Starting from Gaussian noise, it iteratively refines an action chunk conditioned on observation features. 
Receding-horizon execution incorporates updated observations into subsequent predictions, while the generative formulation accommodates multimodal demonstration behaviors. 
The diffusion process operates in action space and does not require a pretrained video-generation backbone. 

\textbf{Octo}~\citep{ghosh2024octo} is a generalist robot policy pretrained on heterogeneous multi-robot demonstrations. 
Modality-specific tokenizers encode observations and task specifications, a Transformer fuses the resulting tokens, and a diffusion action head produces continuous action chunks. 
Tasks can be specified through language instructions or goal images, and the modular design supports adaptation to new observation and action spaces. 
Although Octo uses a pretrained T5 text encoder, its policy backbone is not initialized from a pretrained VLM. 

\textbf{ProgVLA}~\citep{kim2026progvla} combines separate pretrained visual and language encoders with a compact flow-matching action expert. 
Two stages of Perceiver resampling, placed before and after multimodal fusion, compress observation and instruction features into a small set of context tokens. 
Auxiliary progress heads, trained with offline reinforcement-learning objectives, estimate task progress and provide advantage- and success-based weights for the imitation objective. 
This combines representation compression with progress-aware policy training without relying on a pretrained VLM or video-generation backbone.
We only consider it in LIBERO, as it does not currently release code.

\subsection{Real-World}\label{app:real}

\begin{figure}[h]
    \centering
    \includegraphics[width=1.0\textwidth]{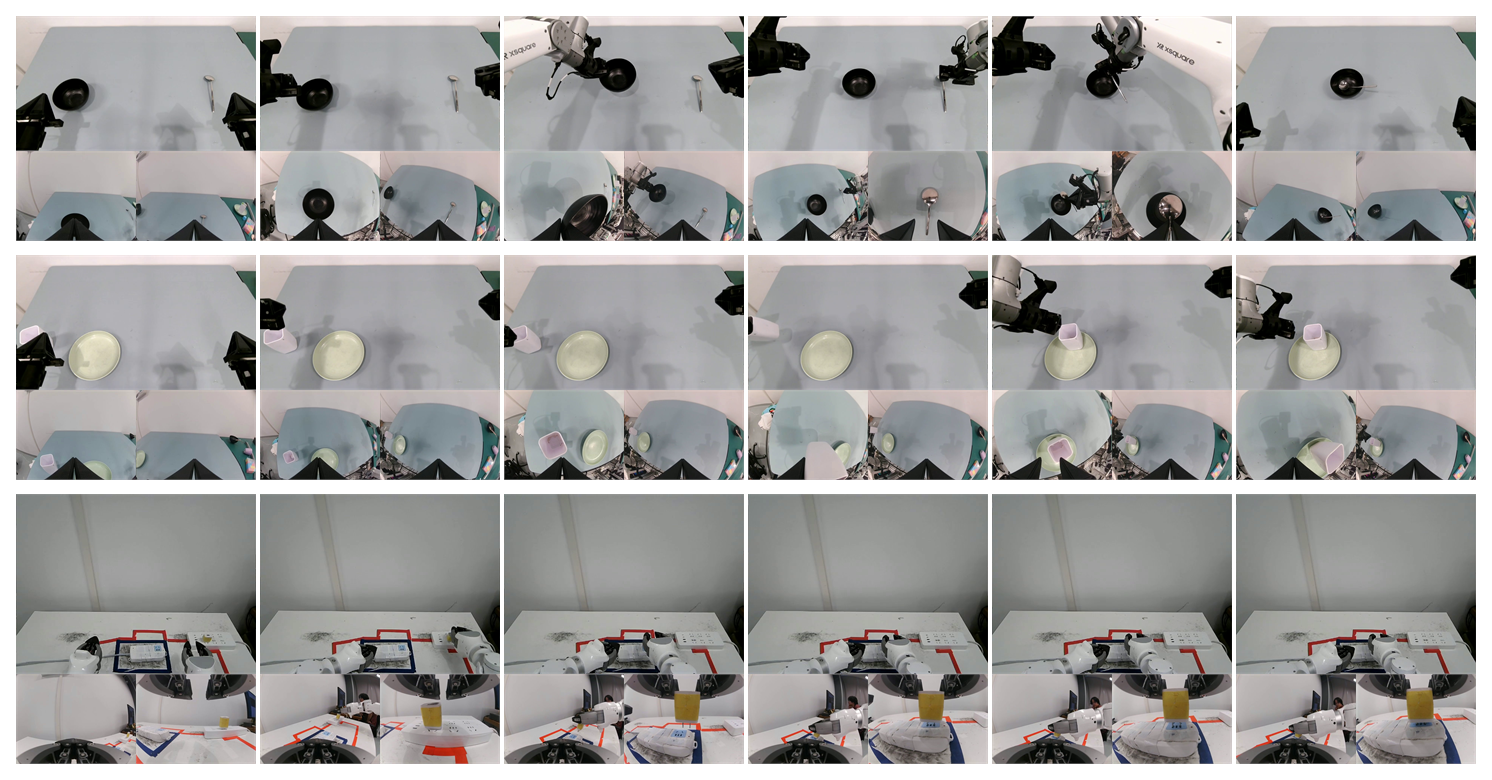}
    \caption{
    \textbf{Example CoRP rollouts on real-world bimanual tasks.}
    Rows show place spoon (top), pick cup (middle), and plug insertion (bottom), with six frames per rollout from left to right in time;
    each frame combines the head-camera view (top) with the two wrist-camera views (bottom).
}
    \label{fig:full_view}
\end{figure}

\paragraph{Robot platform.}
We conduct real-world experiments on two dual-arm robots.
The policy observes the scene through three cameras (Figure~\ref {fig:full_view}): one mounted on the head and one on each wrist.

\paragraph{Implementation Details.}
We make no further architectural changes in real-world experiments.
The main difference lies in the training process.
We first pretrain on diverse data for more than 10,000 hours, covering 5887 different tasks collected in our real-world embodiments. 
We train 400k pretraining steps with a batch size of 4096.
Then we fine-tune the pretrained model on 10 tasks for 30k steps with a batch size of 512.
Both processes adopt exactly same loss function ($\mathcal L_{fm}$ and $\mathcal L_{task}$) and hyper parameters ($\lambda = 0.01$, learning rate $5\times10^{-5}$, and EMA decay $0.999$).

The action space is designed as:
\begin{equation}
\mathbf{a}_t = 
(\mathbf p_l, \mathbf R_l, g_l, \mathbf p_r, \mathbf R_r, g_r) \in \mathbb R^{20},
\end{equation}
where $\mathbf p_l$ and $\mathbf p_r$ denote the positions of the left and right end effectors, respectively, $\mathbf R_l$ and $\mathbf R_r$ denote the first two rows of the rotation matrix~\citep{zhou2020continuityrotationrepresentationsneural} for the left and right end effectors, and $g_l$ and $g_r$ denote the gripper state.
We represent actions as delta states, where each action within an action chunk denotes the relative state difference between the corresponding timestep and the chunk's initial state. 
In particular, for rotational states, we compute the relative rotation on the $\mathrm{SO}(3)$ manifold rather than using a Euclidean representation.

For inference, we use synchronous execution, which is the simplest approach.
Because our model is small enough to require about \textit{20ms} for inference, execution latency is low.
The input instructions for the three tasks are directly \textit{pick\_cup}, \textit{place\_the\_spoon\_inside\_the\_bowl}, and \textit{plug\_insert}, respectively.


\end{document}